\documentclass{article} % For LaTeX2e
\usepackage{iclr2027_conference,times}

\usepackage{amsmath,amsfonts,bm}

\def\eqref#1{equation~\ref{#1}}
\def\1{\bm{1}}

\DeclareMathAlphabet{\mathsfit}{\encodingdefault}{\sfdefault}{m}{sl}
\SetMathAlphabet{\mathsfit}{bold}{\encodingdefault}{\sfdefault}{bx}{n}

\usepackage{hyperref}
\usepackage{url}
\usepackage{multirow}
\usepackage[table,xcdraw]{xcolor}
\usepackage{pifont}
\usepackage{graphicx}
\usepackage{wrapfig}
\usepackage{fontawesome5}
\usepackage[normalem]{ulem}
\useunder{\uline}{\ul}{}

\title{RoPE-Flow: Monocular Vision-based Robot Pose Estimation via 2D Graph-Conditioned 3D Flow Matching}

\iclrfinalcopy

\author{Liangjing Shao$^{1,2}$\And Wanhao Liu$^{2}$\And Zhiwei Fang$^{1}$\And Rulin Zhou$^{1,2}$\And Quanlu Zhang$^{2,3}$\And Changjing Liu$^{1}$\And Beilei Cui$^{1}$\And Yiming Huang$^{1}$\And Hongliang Ren$^{1,2}$\thanks{Corresponding Author} \And  \\
$^{1}$Department of Electronic Engineering, The Chinese University of Hong Kong\\
$^{2}$EACV Research Center, Shenzhen Loop Area Institute\\
$^{3}$Xi'an Jiaotong University
}

\definecolor{first}{HTML}{FBE4E4}
\definecolor{second}{HTML}{FEFFDC}
\definecolor{third}{HTML}{E9FFE5}
\definecolor{green}{HTML}{45be6f}
\definecolor{yellow}{HTML}{d1c54f}
\definecolor{red}{HTML}{ae5b57}

\begin{document}

\maketitle

\begin{figure*}[!ht]
\centering
\includegraphics[width=\textwidth]{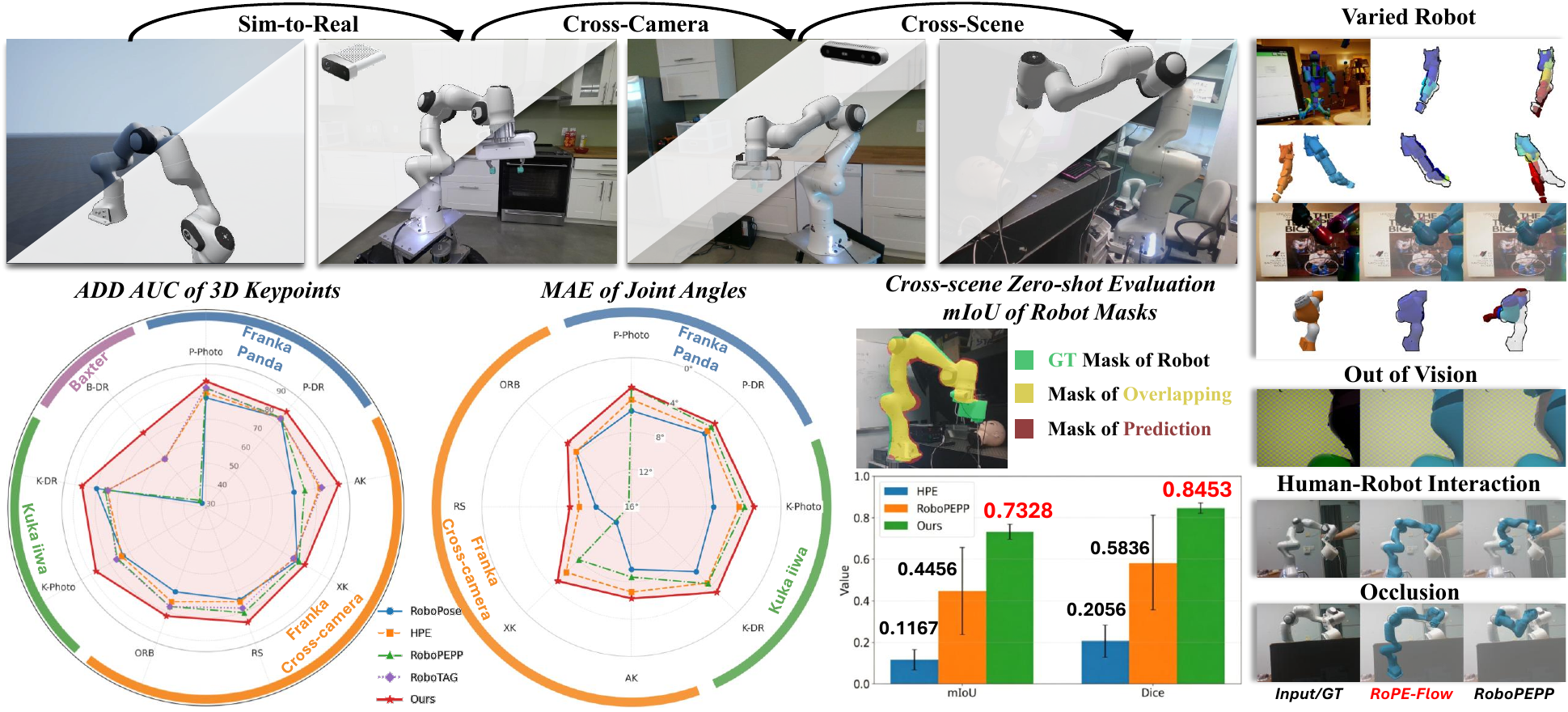}
\caption{Overview of our work. RoPE-Flow can reconstruct \textbf{varied robot} pose \textbf{in diverse scenes} with \textbf{camera-frame 3D keypoints} and \textbf{joint angles} estimation from \textbf{monocular observation of diverse cameras} even with \textbf{human-robot interaction}, \textbf{occlusion} and \textbf{out-of-vision body}.}
\label{fig:teaser}
\end{figure*}

\begin{abstract}
Camera-to-robot pose estimation from a monocular image is significant for vision-based robot navigation and control. However, recovering 3D structural features from 2D observations remains a challenging task. Existing camera-to-robot pose estimation methods either rely on time-consuming iterative rendering or involve sophisticated 2D–3D registration which suffers from projection ambiguity, accumulated errors for distal joints and imbalanced learning for different joints. In this work, we propose \textbf{RoPE-Flow}, an end-to-end flow matching-based framework for camera-to-robot pose estimation. From the predicted 2D heatmaps, we propose a Keypoint-centered Feature Extractor to obtain positional features along with contextual representations from keypoints-centered multi-scale patches. Moreover, we introduce Graph-guided Condition Generator, which embeds a learnable adjacency map to generate condition for flow matching based on correlation among keypoints. Besides, we introduce Bone Length Loss and Ray-Depth Loss for topological consistency and scale stability, with joint-wise weighting to mitigate varied prediction difficulty across joints. The proposed method outperforms state-of-the-art methods for pose estimation across different types of robots with mutiple cameras. Besides, we establish a novel cross-scene benchmark to demonstrate the outstanding performance of RoPE-Flow on zero-shot cross-scene robot pose estimation. Meanwhile, animated visualizations are shown on the website: https://ropeflow.netlify.app/.
\end{abstract}

\section{Introduction}
Accurate pose estimation of articulated robots from monocular RGB is a core capability for vision-based navigation \citep{nav}, human–robot collaboration \citep{hri}, closed-loop manipulation \citep{manip} and multi-robot collaboration \citep{mrc2}. Unlike rigid object pose estimation, robot pose requires recovering the configuration of a kinematic chain whose links are self-occluding and appearance-ambiguous under changing viewpoints and lighting. 

Conventional approaches perform camera-to-robot calibration using externally mounted fiducial markers such as ArUco Markers \citep{arcuo} and AprilTags \citep{apriltag}. The marker’s 2D image coordinates are extracted, while its corresponding 3D position on the robot is computed via forward kinematics. With a collection of across multiple robot configurations, the camera-to-robot transformation can be recovered by Perspective-n-Point (PnP) algorithms. After calibration, both the robot base and camera are assumed to remain stationary. Therefore, calibration can drift from minor impacts or repeated operation, alongside time-varying environmental conditions such as vibration. 

To enable flexible camera-robot calibration for robot interaction in unstructured environments, render-based pipelines \citep{robopose,monose} perform iterative rendering supervised by segmentation masks to directly reconstruct the robot body. However, these methods rely on iterative refinement and require proper initialization for the optimization loop to converge \citep{deepim}. To boost efficiency, end-to-end models \citep{ctr,ctrx,pepp} detect 2D keypoints and recover 3D structure via Perspective-n-Point (PnP), combined with robot forward kinematics (FK) and known camera intrinsics. Still, these approaches derive 3D keypoints from 2D keypoints and joint angles using PnP, with limited robustness due to high sensitivity to correspondence noise. Some recent works \citep{robotag,hpe} adopt 3D models for 2D–3D registration to jointly optimize 2D and 3D representations. Even so, identical 2D projections can correspond to multiple depth configurations, introducing optimization and inference ambiguity. Generative methods offer a promising alternative by modeling the conditional distribution of 3D keypoints rather than predicting single deterministic keypoints estimate\citep{keygen,monose}. Moreover, most existing methods lack zero-shot cross-scene generalization evaluation on real robots under varied backgrounds, they only conduct self-supervised sim-to-real training and tests on single-background real data captured by different cameras from DREAM \citep{dream}.

We propose RoPE-Flow, the first conditioned flow matching-based model for camera-to-robot pose estimation. Method based on flow matching can face two major challenges. \textbf{First}, directly predicting 3D keypoints without PnP and forward kinematics (FK) fails to enforce rigid structural constraints, kinematic priors, and inter-joint relationships. \textbf{Second}, the prediction difficulty varies across different robot joints. To tackle these issues, we make the following contributions:
\begin{enumerate}
    \item We propose a \textbf{Keypoints-centered Feature Extractor} to learn multi-scale contextual representations at keypoint locations together with positional embeddings.
    \item We propose \textbf{Graph-guided Condition Generator} , which provides the condition for 3D flow matching based on keypoint-centric features and the relationships among keypoints.
    \item We introduce \textbf{Bone Length Loss} and \textbf{Ray-Depth Loss} during training to guarantee rigid structural consistency and scale stability, respectively. Meanwhile, \textbf{Joint-wise Weighting} is adopted in loss computation to balance the varied prediction difficulty over joints.
    \item The proposed model outperforms state-of-the-art models across different types of robot with various backgrounds from DREAM dataset. We also construct a novel realistic benchmark for cross-scene robot 3D pose estimation, LabRoPE, to demonstrate outstanding \textbf{zero-shot generalization} of RoPE-Flow.
\end{enumerate}

\section{Related Work}

\subsection{Marker-based Camera-to-Robot Pose Estimation}
The goal of Camera-to-Robot Pose Estimation is to estimate the transformation between a robot’s end-effector and its vision system. Camera-to-Robot Pose Estimation describes the setup where the camera is stationary and positioned away from the robot \citep{tran1,tran2}, in contrast to eye-in-hand setting where the camera is mounted on the robot end-effector \citep{einh}. Traditional camera-to-robot pose estimation approaches attach fiducial markers to the end-effector \citep{arcuo,apriltag}, move the marker to multiple known poses, detect marker locations in captured images, and then solve for the camera-to-robot transformation. While widely applicable, such classic pipeline requires a full offline recalibration whenever the camera-robot relative pose shifts, which becomes cumbersome for real-world deployment. 

\begin{figure}
    \centering
    \includegraphics[width=\linewidth]{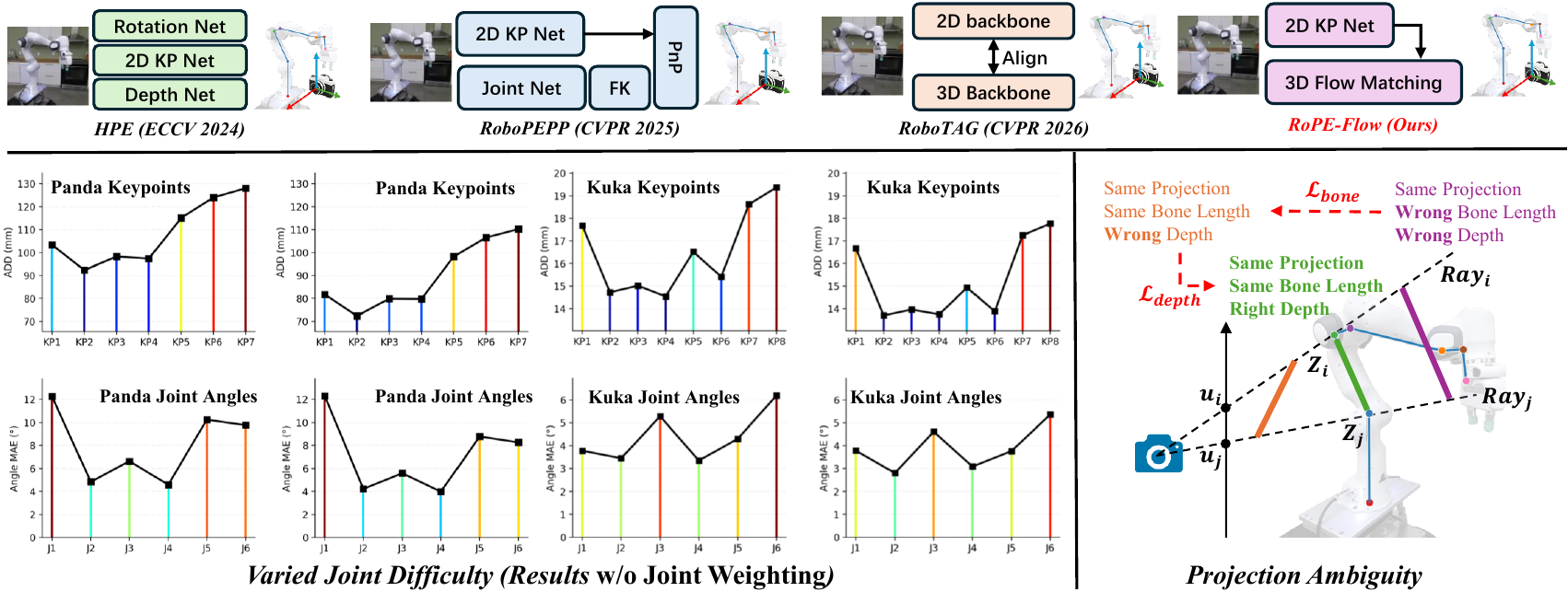}
    \caption{Comparison with previous methods for camera-to-robot estimation (Top Row). Evidence for \textit{Varied Joint Difficulty} is shown based on distribution of prediction errors along the joints and keypoints. Problem for end-to-end 3D keypoints generation: \textit{Projection Ambiguity}, as well as potential solutions in this work is also visualized.}
    \label{issues}
\end{figure}

\subsection{Markerless Camera-to-Robot Pose Estimation}
To enable flexible online camera-to-robot pose estimation for real-world deployment, DREAM \citep{dream} pioneers online markerless hand-eye calibration with a large-scale dataset. RoboPose \citep{robopose} uses iterative rendering supervised by alignment between projected robot models and segmentation masks for pose estimation. Though accurate, its iterative optimization causes heavy computational overhead and slow inference. To resolve this bottleneck, keypoint-based methods are proposed for efficient end-to-end pose estimation. CtRNet \citep{ctr} and CtRNet-X \citep{ctrx} predict 2D keypoints and recover robot poses via PnP given known joint angles. SGTAPose \citep{sgta} exploits temporal cues through a sequential framework. RoboPEPP \citep{pepp} adopts embedding predictive pretraining on masked-joint images to mitigate occlusions, while HPE \citep{hpe} integrates a depth branch to leverage 3D priors. However, all these approaches require a predefined robot ROI before inference. Moreover, pose reconstruction based on PnP and forward kinematics is still sensitive to noise which leads to limited robustness. RoboTAG \citep{robotag} incorporates 3D priors via multi-way 2D–3D alignment, yet it still suffers from occlusions and projection ambiguity. In contrast, our RoPE-Flow directly predicts camera-frame 3D pose from full image and enables uncertainty modeling for occluded joints, dealing with problems from projection ambiguity for direct 3D keypoints generation and varied difficulties for prediction of each joints shown in Fig. \ref{issues}.

\section{Proposed Method}

\begin{figure}
    \centering
    \includegraphics[width=\linewidth]{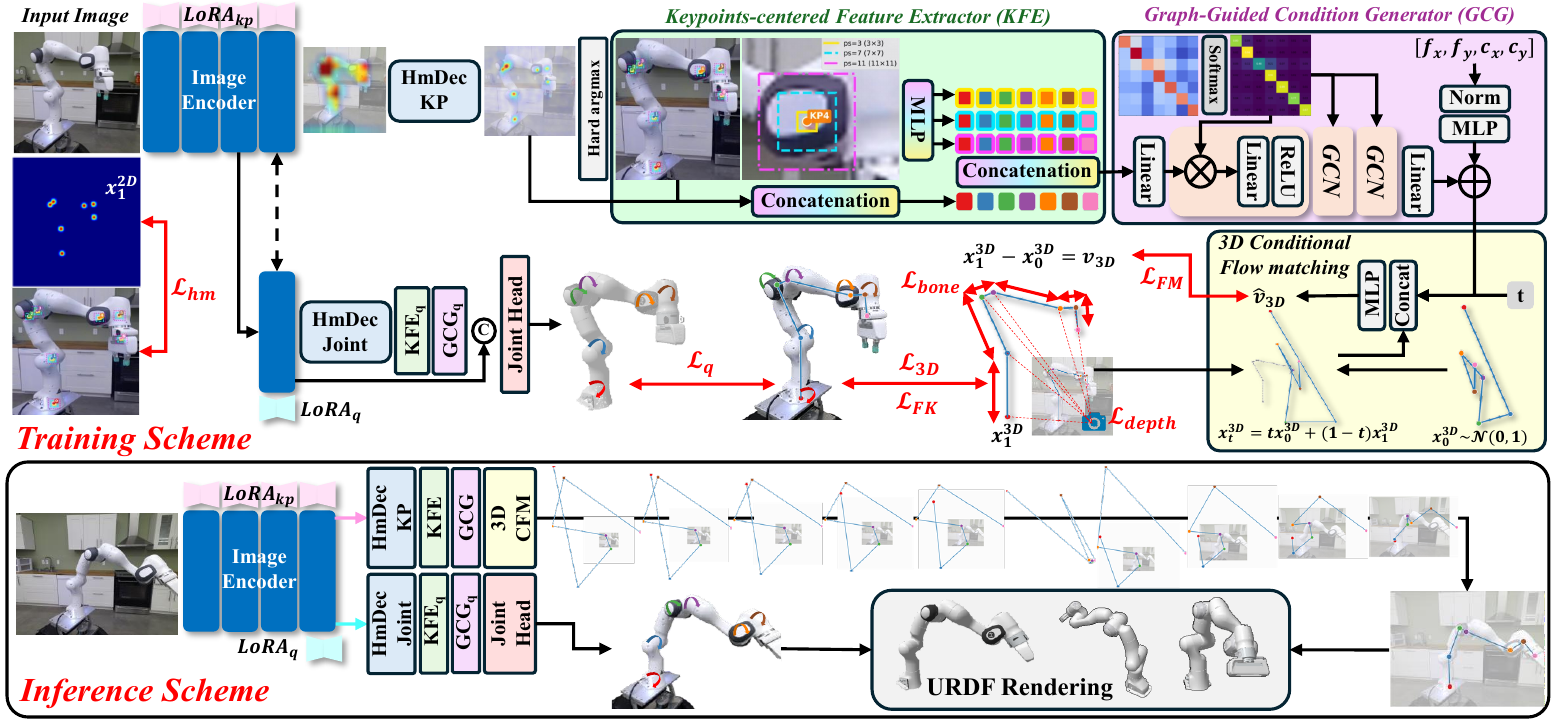}
    \caption{Pipeline of the proposed method. }
    \label{pipe}
\end{figure}

The whole pipeline of RoPE-Flow, shown in Fig. \ref{pipe}, mainly consists of a pretrained image encoder with two groups of LoRA adapters $LoRA_{kp}$ and $LoRA_q$, 2D heatmap decoders $HmDec_{KP}$ and $HmDec_{joint}$, Keypoints-centered Feature Extractor (KFE), Graph-guided Condition Generator (GCG), a Joint Head and a 3D conditional flow matching (3D CFM).

\subsection{2D Keypoints Prediction}
\label{sec:2d}

A pretrained DINOv3 ViT-L/16 produces a spatial feature map $\mathbf{F}\in\mathbb{R}^{C\times h\times w}$ ($C{=}1024$, $h{=}w{=}16$). The backbone remains frozen, while $LoRA_{kp}$ of rank $r{=}32$ and scale $\alpha{=}32$ is injected into projection layers of multi-head self-attention and MLPs of feed forward network. Spatial average pooling of $\mathbf{F}_{img}$ yields a global context $\mathbf{c}_{\mathrm{img}}\in\mathbb{R}^{C}$ that conditions heatmap flow matching during training. Four transposed convolutions followed by a $1{\times}1$ convolution produce per-keypoint belief maps $\mathbf{H} = \mathrm{HmDec}(\mathbf{F}_{img})\in[0,1]^{K\times H\times W}$.
Hard argmax on channel $k$ gives a peak coordinate to denote a position of each keypoint $\mathbf{u}_k=(u_k,v_k)=\arg\max_{(u,v)}
  \mathbf{H}_k(u,v)$, while confidence of each keypoint also comes from the predicted heatmap $s_k=\mathbf{H}_k(u_k,v_k)$.

\subsection{Keypoints-centered Feature Extractor}
Around peak $\mathbf{u}_k$, we bilinearly sample patches $\mathbf{P}_{k,s}$ of sizes $S=\{3,7,11\}$ from $\mathbf{H}_k$.
Each flattened patch is mapped by a two-layer MLP $\phi_s$, then three embeddings are concatenated and fused by an MLP $\psi$:
\begin{equation}
\label{eq:patch}
  \mathbf{p}_k
  =
  \psi\!\Big(
    \big[\phi_3(\mathbf{P}_{k,3});\,
         \phi_7(\mathbf{P}_{k,7});\,
         \phi_{11}(\mathbf{P}_{k,11})\big]
  \Big).
\end{equation}
The final keypoint representation is $\mathbf{f}_k=[\mathbf{u}_k;s_k;\mathbf{p}_k]$, linearly embedded to a latent code $\mathbf{F}_k\in\mathbb{R}^{d_k}$.

\subsection{Graph-guided Condition Generator}
A shared adjacency $\mathbf{A}\in\mathbb{R}^{K\times K}$, initialized as $\mathbf{I}$, is processed by softmax along the row dimension. Given input feature map $\mathbf{F}_{in}$, one graph convolution provides the processed feature map $\mathbf{F}_{out}$ based on connections among keypoints $\mathbf{F}_{out} = \mathrm{softmax}(\mathbf{A})\,\mathbf{F}_{in}\mathbf{W}$.

Gathering keypoints feature $\mathbf{F}\in\mathbb{R}^{K\times d_k}$ will be processed by three graph convolution layers and flattened, linearly projected to keypoint-based condition $\mathbf{c}_{kp}\in\mathbb{R}^{512}$. Meanwhile, normalized intrinsics $\boldsymbol{\kappa}=(f_x/W,\,f_y/H,\,c_x/W,\,c_y/H)$ are mapped by a two-layer MLP whose last layer is zero-initialized, then added residually to form the final condition code $\mathbf{c}$ for flow matching, as Eq. \ref{cam} shows.
\begin{equation}
\label{cam}
  \mathbf{c}
  =
  \mathbf{c}_{\mathrm{gcn}} + \mathrm{CamEnc}(\boldsymbol{\kappa})
  \in\mathbb{R}^{512}.
\end{equation}

\subsection{3D Conditioned Flow Matching}
\label{sec:fm3d}

Camera-frame 3D keypoints are generated under the graph condition $\mathbf{c}$ via conditional flow matching.
Let $\mathbf{x}^{3D}_1=\mathbf{X}^{\mathrm{gt}}\in\mathbb{R}^{K\times 3}$ and $\mathbf{x}^{3D}_0\sim\mathcal{N}(\mathbf{0},\mathbf{I}_{K\times 3})$, the interpolant $\mathbf{x}^{3D}_t$ and target velocity $\mathbf{x}^{3D}_1-\mathbf{x}^{3D}_0$ follow Eq. \ref{ot3d}. A four-layer velocity MLP $v_{3D}$ is utilized to predict velocity from concatenation of intepolant distribution, time token and condition $\big[\mathbf{x}^{3D}_t;\; t;\; \mathbf{c}\big]$.
\begin{equation}
\label{ot3d}
  \mathbf{x}^{3D}_t = (1-t)\mathbf{x}^{3D}_0 + t\mathbf{x}^{3D}_1
\end{equation}

Sampling starts from standard Gaussian noise $\mathbf{x}^{3D}_0\sim\mathcal{N}(\mathbf{0},\mathbf{I}_{K\times 3})$ and applies $N$ Euler steps to generate final predicted 3D Keypoints $\hat{\mathbf{X}}=\mathbf{x}^{3D}_1$, shown as Eq. \ref{ode}
\begin{equation}
\label{ode}
  \mathbf{x}^{3D}_{s+\Delta s}
  =
  \mathbf{x}^{3D}_s + \Delta s\cdot v_{3D}(\mathbf{y}_s,s,\mathbf{c}),
  \quad
  \Delta s = 1/N
\end{equation}

\subsection{Joint Angle Prediction}
A second adapter set $\mathrm{LoRA}_q$ is placed on the last 6 ViT blocks, together with a dedicated 2D keypoint decoder $HmDec_q$ and keypoint-based condition generator $\mathcal{F}^q_{cg}$. Different from 3D keypoints estimation, the condition for joint angle prediction is concatenation of pooled image feature $\mathbf{f}_{\mathrm{img}}^{(q)}$ and keypoints-aware condition $\mathbf{z}=\big[\mathbf{f}_{\mathrm{img}}^{(q)};\,\mathcal{F}^q_{cg}(HmDec_q(\mathbf{f}_{\mathrm{img}}^{(q)}))\big]$.
Following RoboPEPP \citep{pepp}, we use an iterative residual MLP $g_{\phi}$ as Joint Head to perform 4-step iteration of Eq. \ref{jp}, starting from $\bar{\mathbf{q}}^{(0)}=\mathbf{0}$:
\begin{equation}
\label{jp}
  \bar{\mathbf{q}}^{(i+1)}
  =
  \bar{\mathbf{q}}^{(i)}
  +
  g_{\phi}\!\big(
    [\mathbf{z};\,\bar{\mathbf{q}}^{(i)}]
  \big).
\end{equation}

\subsection{Training Losses}
\label{sec:losses}

\subsubsection{Training on Synthetic Data}
\label{sec:syn}

Synthetic supervision has three training stages for 2D keypoints prediction, 3D keypoints prediction and joint angle prediction, respectively.

\paragraph{Stage I: 2D keypoint prediction.}
In stage I, only encoder-decoder architecture for 2D keypoints prediction is trained by $\mathcal{L}_{\mathrm{hm}}$. Ground-truth locations of 2D keypoints $\{u^{gt}_k,v^{gt}_k\}$ are rendered as isotropic Gaussians $\{\mathbf{H}^{\mathrm{gt}}_k\}$. Following \cite{pepp}, we further use CornerNet-style modified focal loss~\citep{law2018cornernet} to supervise 2D keypoints prediction.
% \begin{equation}
% \label{loss1}
%   \mathcal{L}_{\mathrm{hm}}
%   =
%   -\frac{1}{N}
%   \sum_{k,u,v}
%   w_k
%   \begin{cases}
%     (1-\hat{p})^{\alpha}\log\hat{p},
%       & p{=}1, \\[4pt]
%     (1-p)^{\beta}\,\hat{p}^{\alpha}\log(1-\hat{p}),
%       & \text{otherwise},
%   \end{cases}
% \end{equation}
% where $\hat{p}=\mathbf{H}_k(u,v)$, $p=\mathbf{H}^{\mathrm{gt}}_k(u,v)$, and $N$ counts valid pixels.

\paragraph{Stage II: 3D keypoints prediction.}
In Stage II, we keep training Encoder and heatmap decoder with small learning rate, while modules in keypoint-centered feature extractor, graph-guided condition generator and 3D flow matching are mainly optimized. We utilize \textbf{weighted joint-wise velocity loss} of Eq. \ref{fm} to train 3D flow matching.
\begin{equation}
\label{fm}
  \mathcal{L}_{\mathrm{FM}}
  =
  \mathbb{E}_{t,\mathbf{y}_0}
  \sum_{k=1}^{K}
  w_k
  \big\|
    v_{\theta_2}(\mathbf{y}_t,t,\mathbf{c})_k
    -
    (\mathbf{y}_1-\mathbf{y}_0)_k
  \big\|_2^{2}.
\end{equation}
Geometric loss on the predicted 3D keypoints $\hat{\mathbf{X}}$ consists of a \textbf{weighted joint-wise 3D L1} (Eq. \ref{l3d}), \textbf{Bone-Length Loss} (Eq. \ref{bone}), and FK alignment (Eq. \ref{fk}).
\begin{equation}
    \label{l3d}
  \mathcal{L}_{\mathrm{3D}}
  =
  \sum_{k=1}^{K}
  w_k
  \big\|
    \hat{\mathbf{X}}_k - \mathbf{X}^{\mathrm{gt}}_k
  \big\|_1
\end{equation}

\begin{equation}
\label{bone}
  \mathcal{L}_{\mathrm{bone}}
  =
  \frac{1}{|\mathcal{E}|}
  \sum_{(i,j)\in\mathcal{E}}
  \Big|
    \|\hat{\mathbf{X}}_i-\hat{\mathbf{X}}_j\|_2
    -
    \|\mathbf{X}^{\mathrm{gt}}_i-\mathbf{X}^{\mathrm{gt}}_j\|_2
  \Big|
\end{equation}

\begin{equation}
    \label{fk}
  \mathcal{L}_{\mathrm{FK}}
  =
  \big\|
    \hat{\mathbf{X}}_r - \mathrm{FK}(\mathbf{q}^{\mathrm{gt}})
  \big\|_1
\end{equation}
where $\hat{\mathbf{X}}_r=(\mathbf{R}^{\mathrm{gt}})^\top\big(\hat{\mathbf{X}} - \mathbf{t}^{\mathrm{gt}}\big)$ , $\mathrm{FK}(\mathbf{q}^{\mathrm{gt}})$ denotes URDF positions of the seven supervised links and $\mathbf{T}_{c\leftarrow r}^{\mathrm{gt}}$ is the ground-truth extrinsics.
The overall training objective for Stage II is formulated as:
\begin{equation}
  \mathcal{L}_{\mathrm{II}}
  =
  \lambda_{\mathrm{hm}}'\,\mathcal{L}_{\mathrm{hm}}
  +
  \lambda_{\mathrm{FM}}\,\mathcal{L}_{\mathrm{FM}}
  +
  \lambda_{\mathrm{3D}}\,\mathcal{L}_{\mathrm{3D}}
  +
  \lambda_{\mathrm{bone}}\,\mathcal{L}_{\mathrm{bone}}
  +
  \lambda_{\mathrm{FK}}\,\mathcal{L}_{\mathrm{FK}},
\label{l2}
\end{equation}
with
$(\lambda_{\mathrm{hm}}',\lambda_{\mathrm{FM}},\lambda_{\mathrm{3D}},\lambda_{\mathrm{bone}},\lambda_{\mathrm{FK}})
=(0.2,\,1,\,2,\,1,\,1)$.

\paragraph{Stage III: joint-angle prediction head.}
The supervised loss of MSE in normalized joint space Eq. \ref{lj} is utilized to train the pathway for joint angles prediction, with frozen pathway for 3D keypoints prediction.
\begin{equation}
\label{lj}
  \mathcal{L}_{\mathrm{joint}}
  =
  \frac{1}{6}
  \sum_{j=1}^{6}
  \big(
    \bar{q}_j - \bar{q}_j^{\mathrm{gt}}
  \big)^{2}.
\end{equation}

\subsubsection{Sim-to-Real Fine-tuning}
\label{sec:s2r}

We firstly finetune the pathway for 3D keypoints prediction except keypoints decoder on realistic data.  3D keypoints $\hat{\mathbf{X}}_{b,k}$ obtained from the model trained on synthetic data are mapped back to original-image pixels by $\hat{u}_{b,k}=\pi(\hat{\mathbf{X}}_{b,k};\mathbf{K})$ with predicted 3D keypoints $\hat{\mathbf{X}}$.
The reprojection residual of the sample $b$ is $r_b=\frac{1}{K}\sum_{k=1}^{K}\big\|\hat{\mathbf{u}}_{b,k} - \mathbf{u}_{b,k}\big\|_1$, where $\mathbf{u}$ denotes heatmap peaks predicted by synthetic-domain model. Samples with $r_b<\tau_{\mathrm{px}}$ ($\tau_{\mathrm{px}}{=}100$) form the inlier set $\mathcal{V}$, otherwise the lowest $\rho{=}50\%$ of finite residuals are kept.
The Smooth-L1 reprojection loss is defined as:
\begin{equation}
\label{eq:reproj}
  \mathcal{L}_{\mathrm{reproj}}
  =
  \frac{1}{|\mathcal{V}|}
  \sum_{b\in\mathcal{V}}
  \mathrm{SmoothL1}\!\big(
    \pi(\hat{\mathbf{X}}_b;\mathbf{K}),\,\mathbf{u}_b
  \big).
\end{equation}
We formulate bone length loss with adjacent supervised-link distances $\ell_{ij}$ at $\mathbf{q}{=}\mathbf{0}$ (pose-invariant):
\begin{equation}
\label{eq:bone-urdf}
  \mathcal{L}_{\mathrm{bone}}^{\mathrm{S2R}}
  =
  \frac{1}{|\mathcal{E}|}
  \sum_{(i,j)\in\mathcal{E}}
  \Big|
    \|\hat{\mathbf{X}}_i-\hat{\mathbf{X}}_j\|_2 - \ell_{ij}
  \Big|.
\end{equation}
For each keypoint we take the predicted depth $z_k=\hat{X}_{k,z}$ and lift the detached 2D peak $\mathbf{u}_k$ along the corresponding camera ray $\mathbf{d}_k=\mathbf{K}^{-1}[u_k,v_k,1]^\top$, yielding a ray-consistent point $\tilde{\mathbf{X}}_k=z_k\mathbf{d}_k$. We then align prediction $\hat{\mathbf{X}}$ to this lift by Eq. \ref{depth} with $z_{\min}{=}0.05\,\mathrm{m}$.
\begin{equation}
\label{depth}
  \mathcal{L}_{\mathrm{depth}}
  =
  \mathrm{SmoothL1}\!\big(\hat{\mathbf{X}},\,\tilde{\mathbf{X}}\big)
  +
  \frac{1}{|\mathcal{E}|}
  \sum_{(i,j)\in\mathcal{E}}
  \Big|
    \|\tilde{\mathbf{X}}_i-\tilde{\mathbf{X}}_j\|_2 - \ell_{ij}
  \Big|
  +
  \frac{1}{K}
  \sum_{k}
  \mathrm{ReLU}(z_{\min}-z_k),
\end{equation}
The final sim-to-real training loss is formulated as Eq. \ref{s2r}, where $\lambda_{\mathrm{bone}}^{\mathrm{S2R}}=20,\lambda_{\mathrm{depth}}=40$
\begin{equation}
\label{s2r}
  \mathcal{L}^{kp}_{\mathrm{S2R}}
  =
  \mathcal{L}_{\mathrm{reproj}}
  +
  \lambda_{\mathrm{bone}}^{\mathrm{S2R}}\,
  \mathcal{L}_{\mathrm{bone}}^{\mathrm{S2R}}
  +
  \lambda_{\mathrm{depth}}\,\mathcal{L}_{\mathrm{depth}}
\end{equation}

Additionally, we adapt pathway for joint angle prediction following self-supervised sim-to-real finetuning in \cite{pepp}, based on 2D keypoints reprojected from 3D keypoints generated by finetuned 3D keypoints prediction pathway.

\section{Experiments and Results}

\subsection{Implementation Details}
Following \citep{pepp,robotag}, we train and evaluate our method on the DREAM dataset \citep{dream}. The model is separately trained on the synthetic DR split of Franka, Kuka and Baxter. Evaluation is conducted on DR and Photos splits for Panda and Kuka, and the DR split for Baxter. For Franka, we perform sim-to-real finetuning on Azure (AK) data using DR pretrained weights, then train and test on Kinect360 (XK), RealSense (RS) and ORB data based on weights for AK.

For synthetic training, we apply random $256{\times}256$ cropping, color jitter, Gaussian blur and horizontal flipping. Staged training runs 100, 300 and 300 epochs with batch size 32 for Stage I--III under the OneCycle schedule. Joint-wise weights are [1.0, 1.0, 1.0, 1.0, 1.5, 2.0, 2.0], [1.0, 1.0, 1.0, 1.0, 1.5, 2.0, 2.0, 2.5] and [1.0, 1.0, 1.0, 1.0, 1.0, 1.2, 1.2, 1.5, 1.5, 2.0, 2.0, 2.5, 2.5, 3.0, 3.0, 3.5, 3.5] for Panda, Kuka and Baxter, respectively. Sim-to-real finetuning takes 20 epochs with batch size $16$ and no augmentation. All experiments are conducted on one NVIDIA H100 GPU.

\begin{table}[!ht]
\centering
\begin{tabular}{c|cccccc|cc|c|c}
\hline
                                        & 
\multicolumn{6}{c|}{\cellcolor[HTML]{EFEFEF}\textbf{Panda}}                                                                                                                             & \multicolumn{2}{c|}{\cellcolor[HTML]{EFEFEF}\textbf{Kuka}}  & \cellcolor[HTML]{EFEFEF}\textbf{Baxter}
 &                                \\
\multirow{-2}{*}{\textbf{Methods}}      & \textbf{Photo}               & \textbf{DR}                  & \textbf{AK}                  & \textbf{XK}                  & \textbf{RS}                  & \textbf{ORB}                 & \textbf{Photo}               & \textbf{DR}                  & \textbf{DR}                             & \multirow{-2}{*}{\textbf{Avg}} \\ \hline
{
\color[HTML]{C0C0C0} \textbf{DREAM-F*}} & {\color[HTML]{C0C0C0} 79.5}  & {\color[HTML]{C0C0C0} 81.3}  & {\color[HTML]{C0C0C0} 68.9}  & {\color[HTML]{C0C0C0} 24.4}  & {\color[HTML]{C0C0C0} 76.1}  & {\color[HTML]{C0C0C0} 61.9}  & {\color[HTML]{C0C0C0} /}     & {\color[HTML]{C0C0C0} /}     & {\color[HTML]{C0C0C0} /}                & {\color[HTML]{C0C0C0}
 /}       \\
{
\color[HTML]{C0C0C0} \textbf{DREAM-Q*}} & {\color[HTML]{C0C0C0} 74.3}  & {\color[HTML]{C0C0C0} 77.8}  & {\color[HTML]{C0C0C0} 52.4}  & {\color[HTML]{C0C0C0} 37.5}  & {\color[HTML]{C0C0C0} 78.0}  & {\color[HTML]{C0C0C0} 57.0}  & {\color[HTML]{C0C0C0} /}     & {\color[HTML]{C0C0C0} /}     & {\color[HTML]{C0C0C0} 75.5}             & {\color[HTML]{C0C0C0}
 /}       \\
{
\color[HTML]{C0C0C0} \textbf{DREAM-H*}} & {\color[HTML]{C0C0C0} 81.1}  & {\color[HTML]{C0C0C0} 82.9}  & {\color[HTML]{C0C0C0} 60.5}  & {\color[HTML]{C0C0C0} 64.0}  & {\color[HTML]{C0C0C0} 78.8}  & {\color[HTML]{C0C0C0} 69.1}  & {\color[HTML]{C0C0C0} 72.1}  & {\color[HTML]{C0C0C0} 73.3}  & {\color[HTML]{C0C0C0} /}                & {\color[HTML]{C0C0C0}
 /}       \\
\textbf{RoboPose$\dagger$}                       & 79.7                         & \cellcolor[HTML]{E9FFE5}82.9 & 70.4                         & \cellcolor[HTML]{E9FFE5}77.6 & 74.3                         & 70.4                         & 73.2                         & \cellcolor
[HTML]{FEFFDC}80.2 & 32.7                                    & 71.3                           \\
\textbf{HPE$\ddagger$}                            & 82.0                         & 82.7                         & \cellcolor[HTML]{E9FFE5}82.2 & 76.0                         & 75.2                         & \cellcolor[HTML]{E9FFE5}75.2 & 73.9                         & 75.1                         & \cellcolor[HTML]{FEFFDC}58.8            & \cellcolor
[HTML]{E9FFE5}75.7   \\
\textbf{HPE$\dagger$}
                           & 40.7                         & 41.4                         & 66.7                         & /                            & 49.1                         & 51.6                         & 56.7                         & 56.2                         & 9.8                                     & /                              \\
\textbf{RoboPEPP$\dagger$}                       & \cellcolor[HTML]{E9FFE5}84.1 & \cellcolor[HTML]{FEFFDC}83.0 & 75.3                         & \cellcolor[HTML]{FEFFDC}78.5 & \cellcolor[HTML]{FEFFDC}80.5 & \cellcolor[HTML]{FEFFDC}77.5 & \cellcolor[HTML]{E9FFE5}76.1 & \cellcolor[HTML]{E9FFE5}76.2 & \cellcolor
[HTML]{E9FFE5}34.4            & 74.0                           \\
\textbf{RoboTAG}                        & \cellcolor[HTML]{FEFFDC}84.3 & 82.5                         & \cellcolor[HTML]{FEFFDC}83.1 & 75.7                         & \cellcolor[HTML]{E9FFE5}78.3 & \cellcolor[HTML]{FEFFDC}77.5 & \cellcolor[HTML]{FEFFDC}76.6 & 75.0                         & \cellcolor[HTML]{FEFFDC}58.8            & \cellcolor
[HTML]{FEFFDC}76.9   \\
\textbf{RoPE-Flow}                      & \cellcolor[HTML]{FBE4E4}87.1 & \cellcolor[HTML]{FBE4E4}86.6 & \cellcolor[HTML]{FBE4E4}90.9 & \cellcolor[HTML]{FBE4E4}81.6 & \cellcolor[HTML]{FBE4E4}85.1 & \cellcolor[HTML]{FBE4E4}82.1 & \cellcolor[HTML]{FBE4E4}87.3 & \cellcolor[HTML]{FBE4E4}86.9 & \cellcolor[HTML]{FBE4E4}74.1            & \cellcolor[HTML]{FBE4E4}84.6   \\ \hline
\end{tabular}
\caption{Comparison of robot pose estimation on DREAM dataset using AUC on the ADD metric. *DREAM and its variants need ground-truth joint state parameters. $\ddagger$ Results with ground-truth bounding box. $\dagger$ Results with predicted bounding box. \colorbox{first}{\textbf{The Best}}, \colorbox{second}{\textbf{Second Best}} and \colorbox{third}{\textbf{Third Best}} results are marked by corresponding colors. Curves of ADD can be found in \textit{Appendix} \ref{cadd}.}
\label{add}
\end{table}

\subsection{Camera-frame Pose Estimation}
We compare RoPE-Flow with prior 3D keypoints prediction methods on the DREAM dataset, using the Area Under Curve (AUC) of Average Distance Distribution (ADD) as in previous works \citep{pepp,robotag}. Table \ref{add} shows our method surpasses SOTA approaches on all splits across different robots and cameras in DREAM. We also adopt Mean Average Distance between predicted and ground-truth camera-frame joint positions for evaluation (Table \ref{mad}). RoPE-Flow yields the most accurate joint positions without joint priors and achieves competitive results against methods using known joint states. Qualitative comparisons of 3D skeletons reconstructed from predicted keypoints are provided in the \textit{Appendix} \ref{qkp}.

\begin{table}[!ht]
\renewcommand{\arraystretch}{1.1}
\centering
\begin{tabular}{c|cccccc|cc|c}
\hline
                                         & 
\multicolumn{6}{c|}{\cellcolor[HTML]{EFEFEF}\textbf{Panda}}                                                                                                                                                                               & \multicolumn{2}{c|}{\cellcolor[HTML]{EFEFEF}\textbf{Kuka}}                    & \cellcolor[HTML]{EFEFEF}\textbf{Baxter}
 \\
\multirow{-2}{*}{\textbf{Methods}}       & \textbf{Photo}                        & \textbf{DR}                         & \textbf{AK}                          & \textbf{XK}                          & \textbf{RS}                           & \textbf{ORB}                          & \textbf{Photo}                        & \textbf{DR}                           & \textbf{DR}                             \\ \hline
{
\color[HTML]{C0C0C0} \textbf{DREAM-F}}  & {\color[HTML]{C0C0C0} 785.8}          & {\color[HTML]{C0C0C0} 552.0}        & {\color[HTML]{C0C0C0} $>$1k}         & {\color[HTML]{C0C0C0} $>$1k}         & {\color[HTML]{C0C0C0} $>$1k}          & {\color[HTML]{C0C0C0} $>$1k}          & {\color[HTML]{C0C0C0} /}              & {\color[HTML]{C0C0C0} /}              & {\color[HTML]{C0C0C0}
 /}                \\
{
\color[HTML]{C0C0C0} \textbf{DREAM-Q}}  & {\color[HTML]{C0C0C0} 569.0}          & {\color[HTML]{C0C0C0} 359.0}        & {\color[HTML]{C0C0C0} $>$1k}         & {\color[HTML]{C0C0C0} $>$1k}         & {\color[HTML]{C0C0C0} 27.2}           & {\color[HTML]{C0C0C0} $>$1k}          & {\color[HTML]{C0C0C0} /}              & {\color[HTML]{C0C0C0} /}              & {\color[HTML]{C0C0C0}
 59.0}             \\
{
\color[HTML]{C0C0C0} \textbf{DREAM-H}}  & {\color[HTML]{C0C0C0} 518.8}          & {\color[HTML]{C0C0C0} 325.6}        & {\color[HTML]{C0C0C0} 56.5}          & {\color[HTML]{C0C0C0} $>$1k}         & {\color[HTML]{C0C0C0} 23.6}           & {\color[HTML]{C0C0C0} $>$1k}          & {\color[HTML]{C0C0C0} $>$1k}          & {\color[HTML]{C0C0C0} $>$1k}          & {\color[HTML]{C0C0C0}
 /}                \\
{
\color[HTML]{C0C0C0} \textbf{RoboPose}} & {\color[HTML]{C0C0C0} 21.5}           & {\color[HTML]{C0C0C0} \textbf{7.9}} & {\color[HTML]{C0C0C0} 24.4}          & {\color[HTML]{C0C0C0} \textbf{14.0}} & {\color[HTML]{C0C0C0} 23.1}           & {\color[HTML]{C0C0C0} {\ul 19.4}}     & {\color[HTML]{C0C0C0} 33.5}           & {\color[HTML]{C0C0C0} {\ul 19.2}}     & {\color[HTML]{C0C0C0} \textbf{15.5}}
    \\
\textbf{RoboPose}
                        & 21.4                                  & 32.2                                & 34.3                                 & 22.3                                 & 26.0                                  & 30.1                                  & 89.0                                  & 34.2                                  & 87.9                                    \\
\textbf{HPE}                             & \cellcolor[HTML]{FEFFDC}18.0          & \cellcolor[HTML]{FEFFDC}18.4        & \cellcolor[HTML]{FEFFDC}{\ul 18.9}   & 24.0                                 & 24.8                                  & \cellcolor[HTML]{FEFFDC}24.8          & \cellcolor[HTML]{FEFFDC}{\ul 29.0}    & \cellcolor[HTML]{FEFFDC}28.0          & \cellcolor
[HTML]{FEFFDC}45.0            \\
\textbf{RoboPEPP}                        & 29.8                                  & 49.5                                & 29.0                                 & \cellcolor[HTML]{FEFFDC}22.0         & \cellcolor[HTML]{FEFFDC}{\ul 23.0}
    & 27.0                                  & 72.4                                  & 84.2                                  & 157.5                                   \\
\textbf{Ours}                            & \cellcolor[HTML]{FBE4E4}\textbf{12.9} & \cellcolor[HTML]{FBE4E4}{\ul 13.8}  & \cellcolor[HTML]{FBE4E4}\textbf{9.1} & \cellcolor[HTML]{FBE4E4}{\ul 18.6}   & \cellcolor[HTML]{FBE4E4}\textbf{14.9} & \cellcolor[HTML]{FBE4E4}\textbf{18.4} & \cellcolor[HTML]{FBE4E4}\textbf{13.0} & \cellcolor[HTML]{FBE4E4}\textbf{14.5} & \cellcolor[HTML]{FBE4E4}{\ul 30.6}      \\ \hline
\end{tabular}
\caption{Comparison of Mean Average Distance (mm). \textbf{The best results} among all methods (including {\color[HTML]{C0C0C0} methods with given joint states}) are in bold, while \underline{the second best one} is underlined. \colorbox{first}{\textbf{The Best}} and \colorbox{second}{\textbf{Second Best}} results among methods \textit{with no need of joint state parameters} are marked by corresponding colors.}
\label{mad}
\end{table}

\subsection{Joint Angle Estimation}

\begin{table}[!ht]
\centering
\begin{tabular}{c|cc|cc|cccc}
\hline
                                   & \multicolumn{2}{c|}{\cellcolor[HTML]{EFEFEF}\textbf{Panda}} & \multicolumn{2}{c|}{\cellcolor[HTML]{EFEFEF}\textbf{Kuka}} & \multicolumn{4}{c}{\cellcolor[HTML]{EFEFEF}\textbf{Panda}}                                                             \\
\multirow{-2}{*}{\textbf{Methods}} & \textbf{Photo}               & \textbf{DR}                  & \textbf{Photo}               & \textbf{DR}                 & \textbf{AK}                 & \textbf{XK}                 & \textbf{RS}                  & \textbf{ORB}                \\ \hline
\textbf{RoboPose}                  & 5.7                          & 4.9                          & 7.2                          & 6.2                         & 9.3                         & 13.7                        & 12.2                         & 7.6                         \\
\textbf{HPE}                       & \cellcolor[HTML]{FEFFDC}4.5  & 4.5                          & 4.5                          & 4.5                         & \cellcolor[HTML]{FEFFDC}6.9 & \cellcolor[HTML]{FEFFDC}6.1 & \cellcolor[HTML]{FEFFDC}10.4 & \cellcolor[HTML]{FEFFDC}7.6 \\
\textbf{RoboPEPP}                  & \cellcolor[HTML]{FBE4E4}3.2  & \cellcolor[HTML]{FEFFDC}3.9  & \cellcolor[HTML]{FEFFDC}3.9  & \cellcolor[HTML]{FEFFDC}4.4 & 8.5                         & 8.0                         & 15.7                         & 15.6                        \\
\textbf{Ours}                      & \cellcolor[HTML]{FBE4E4}3.2  & \cellcolor[HTML]{FBE4E4}3.4  & \cellcolor[HTML]{FBE4E4}2.9  & \cellcolor[HTML]{FBE4E4}3.1 & \cellcolor[HTML]{FBE4E4}6.2 & \cellcolor[HTML]{FBE4E4}4.8 & \cellcolor[HTML]{FBE4E4}9.4  & \cellcolor[HTML]{FBE4E4}6.3 \\ \hline
\end{tabular}
\caption{Comparison of joint angle estimation results using average MAE (°) of joint angles. Detailed MAE on each joint can be found in \textit{Appendix} \ref{jmae}. \colorbox{first}{\textbf{The Best}} and \colorbox{second}{\textbf{Second Best}} results are marked by corresponding colors.}
\label{cq}
\end{table}

\begin{figure}[!ht]
    \centering
    \includegraphics[width=\linewidth]{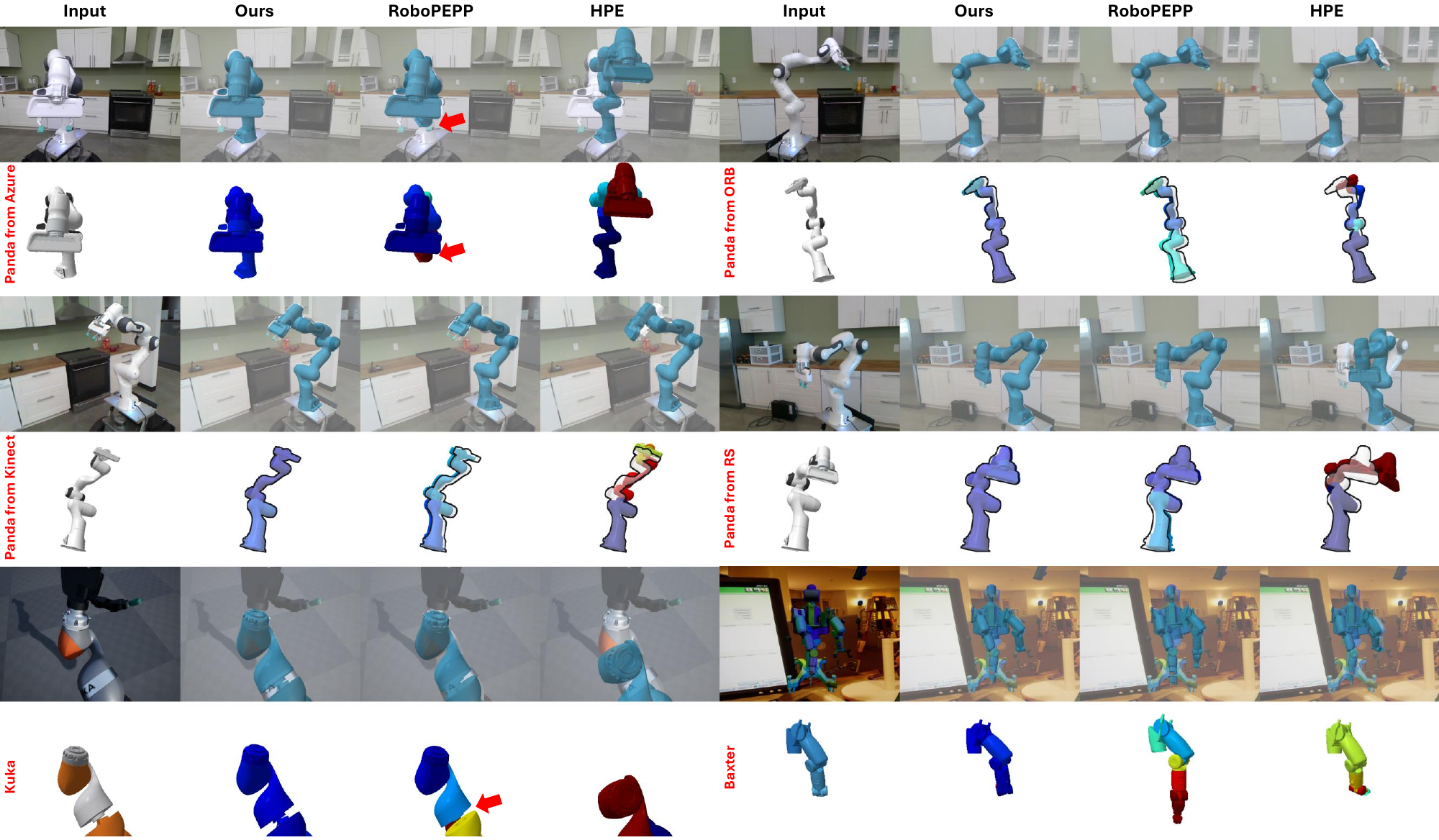}
    \caption{Qualitative comparison on URDF-based reconstructed models. More qualitative results can be found in \textit{Appendix} \ref{sdream}. \textbf{Results on videos} can be found in \textit{Supplementary Material}.}
    \label{mesh}
\end{figure}

We compare our method with RoboPose, HPE and RoboPEPP for joint angle prediction on the Panda and Kuka splits of the DREAM dataset, using the mean average error (MAE) of all joint angles. Since the last Franka joint’s rotation aligns with its corresponding 3D keypoint axis, we only compute errors for the first six joints \citep{pepp}. Quantitative results are listed in Table \ref{cq}, and per-joint details are given in the Appendix. Even though HPE and RoboPEPP rely on ground truth or preprocessed bounding boxes, RoPE-Flow achieves the best joint angle accuracy on both Franka and Kuka. We reconstruct robot 3D models from predicted joint angles and 3D keypoints using provided URDFs for qualitative evaluation. As shown in Fig. \ref{mesh}, our method yields more accurate robot models than prior SOTA approaches.

\subsection{Zero-shot Cross-scene Estimation}

\begin{wraptable}{r}{8cm}
\centering
\begin{tabular}{c|cc}
\hline
\textbf{Methods}  & \textbf{mIoU}           & \textbf{Dice}           \\ \hline
\textbf{HPE}      & 0.1167 ± 0.049          & 0.2056 ± 0.077          \\
\textbf{RoboPEPP} & 0.4456 ± 0.211          & 0.5836 ± 0.229          \\
\textbf{Ours}     & \textbf{0.7328 ± 0.036} & \textbf{0.8453 ± 0.024} \\ \hline
\end{tabular}
\caption{Quantitative comparison on zero-shot cross-scene dataset.}
\label{0comp}
\end{wraptable}

\begin{figure}[!ht]
    \centering
    \includegraphics[width=\linewidth]{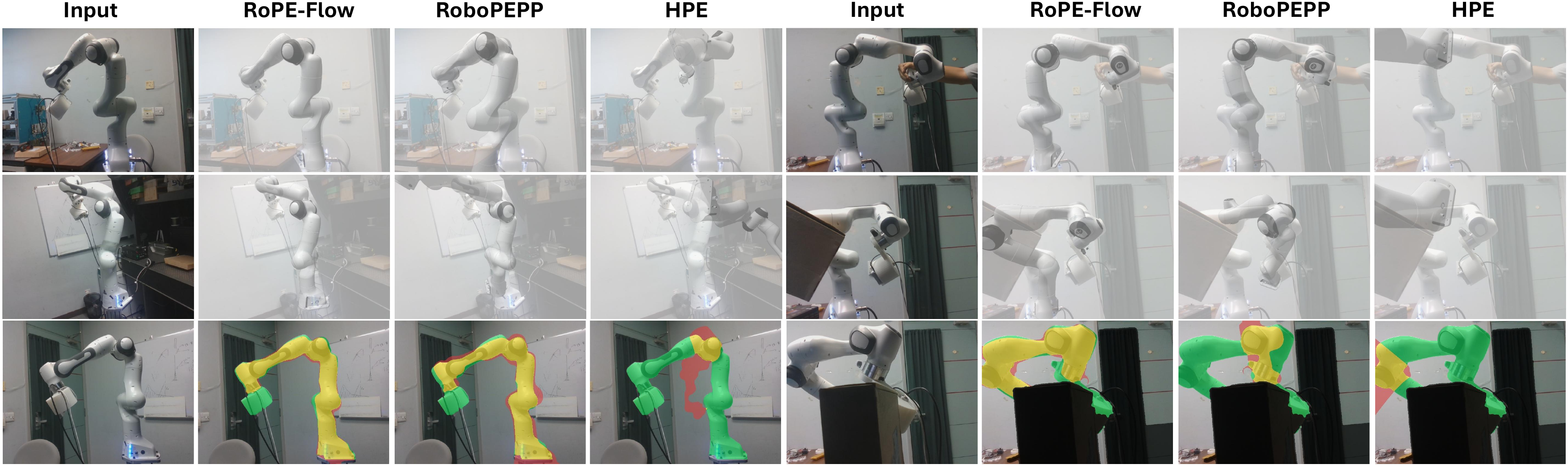}
    \caption{Qualitative comparison for zero-shot cross-scene samples. At last row, we use green mask to denote the ground truth, while yellow mask represents overlapping area and red mask is the mask of reconstructed robot. More examples can be found in \textit{Appendix} \ref{srose}.}
    \label{q3}
\end{figure}

We build LabRoPE (details in Appendix \ref{rope}), a novel zero-shot benchmark containing 50 samples captured by RealSense from diverse views, to complement DREAM’s realistic single-background scenes. Inspired by RoboPose \citep{robopose}, we adopt mIoU and Dice coefficient between robot masks of input image and masks from reconstructed 3D models for zero-shot evaluation. We compare RoPE-Flow with two publicly available SOTA methods, RoboPEPP and HPE. Table \ref{0comp} verifies our method’s strong zero-shot generalization, and qualitative results are presented in Fig. \ref{q3}.

\begin{table}[]
\centering
\begin{tabular}{ccccccccccc}
\hline
\multicolumn{1}{c|}{}                                     & \multicolumn{6}{c|}{\cellcolor[HTML]{C0C0C0}\textbf{Panda}}                                                                                                                                                                                                                             & \multicolumn{2}{c|}{\cellcolor[HTML]{C0C0C0}\textbf{Kuka}}                                         & \multicolumn{1}{c|}{\cellcolor[HTML]{C0C0C0}\textbf{Baxter}} &                                       \\
\multicolumn{1}{c|}{\multirow{-2}{*}{\textbf{Ablations}}} & \textbf{Photo}                        & \multicolumn{1}{c|}{\textbf{DR}}                           & \textbf{AK}                           & \textbf{XK}                           & \textbf{RS}                           & \multicolumn{1}{c|}{\textbf{ORB}}                          & \textbf{Photo}                        & \multicolumn{1}{c|}{\textbf{DR}}                           & \multicolumn{1}{c|}{\textbf{DR}}                             & \multirow{-2}{*}{\textbf{Avg}}        \\ \hline
\multicolumn{1}{c|}{\textbf{Full Model}}                  & \cellcolor[HTML]{FFE8E8}\textbf{87.1} & \multicolumn{1}{c|}{\cellcolor[HTML]{FEFFBB}\textbf{86.6}} & \cellcolor[HTML]{FFE8E8}\textbf{90.9} & \cellcolor[HTML]{FFE8E8}\textbf{81.6} & \cellcolor[HTML]{FFE8E8}\textbf{85.1} & \multicolumn{1}{c|}{\cellcolor[HTML]{FFE8E8}\textbf{82.1}} & \cellcolor[HTML]{FEFFBB}\textbf{87.3} & \multicolumn{1}{c|}{\cellcolor[HTML]{FEFFBB}\textbf{86.9}} & \multicolumn{1}{c|}{\cellcolor[HTML]{FEFFBB}\textbf{74.1}}   & \cellcolor[HTML]{FFE8E8}\textbf{84.6} \\ \hline
\multicolumn{11}{c}{\cellcolor[HTML]{EFEFEF}\textit{\textbf{Keypoints-centered Feature Extraction}}}                                                                                                                                                                                                                                                                                                                                                                                                                                                            \\
\multicolumn{1}{c|}{\textbf{3$\times$3 Patch}}            & 86.0                                  & \multicolumn{1}{c|}{85.8}                                  & 89.3                                  & \cellcolor[HTML]{FEFFBB}81.4          & \cellcolor[HTML]{D3FFDA}84.6          & \multicolumn{1}{c|}{76.4}                                  & 86.8                                  & \multicolumn{1}{c|}{86.2}                                  & \multicolumn{1}{c|}{70.8}                                    & \cellcolor[HTML]{D3FFDA}83.0          \\
\multicolumn{1}{c|}{\textbf{7$\times$7 Patch}}            & \cellcolor[HTML]{D3FFDA}86.6          & \multicolumn{1}{c|}{86.3}                                  & \cellcolor[HTML]{FEFFBB}90.2          & \cellcolor[HTML]{D3FFDA}79.4          & \cellcolor[HTML]{D3FFDA}84.6          & \multicolumn{1}{c|}{79.4}                                  & 86.9                                  & \multicolumn{1}{c|}{86.5}                                  & \multicolumn{1}{c|}{69.9}                                    & \cellcolor[HTML]{FEFFBB}83.3          \\
\multicolumn{1}{c|}{\textbf{11$\times$11 Patch}}          & \cellcolor[HTML]{FEFFBB}87.0          & \multicolumn{1}{c|}{\cellcolor[HTML]{FFE8E8}86.7}          & 89.0                                  & 75.5                                  & 77.2                                  & \multicolumn{1}{c|}{79.7}                                  & \cellcolor[HTML]{FFE8E8}87.6          & \multicolumn{1}{c|}{\cellcolor[HTML]{FFE8E8}87.2}          & \multicolumn{1}{c|}{\cellcolor[HTML]{FFE8E8}74.2}            & 82.7                                  \\
\multicolumn{1}{c|}{\textbf{w/o Patch}}                   & 83.2                                  & \multicolumn{1}{c|}{83.2}                                  & 89.3                                  & 78.2                                  & 83.9                                  & \multicolumn{1}{c|}{\cellcolor[HTML]{FEFFBB}80.5}          & 84.8                                  & \multicolumn{1}{c|}{84.2}                                  & \multicolumn{1}{c|}{73.3}                                    & 82.3                                  \\
\multicolumn{1}{c|}{\textbf{w/o Position}}                & 39.6                                  & \multicolumn{1}{c|}{38.2}                                  & 0.0                                   & 0.0                                   & 0.0                                   & \multicolumn{1}{c|}{\cellcolor[HTML]{FFFFFF}0.0}           & 45.4                                  & \multicolumn{1}{c|}{45.7}                                  & \multicolumn{1}{c|}{29.7}                                    & 22.1                                  \\ \hline
\multicolumn{11}{c}{\cellcolor[HTML]{EFEFEF}\textit{\textbf{2D Graph-guided  3D Flow Matching}}}                                                                                                                                                                                                                                                                                                                                                                                                                                                                \\
\multicolumn{1}{c|}{\textbf{w/o CFM}}                     & 83.4                                  & \multicolumn{1}{c|}{83.2}                                  & 88.8                                  & 75.6                                  & 84.2                                  & \multicolumn{1}{c|}{79.0}                                  & 67.2                                  & \multicolumn{1}{c|}{63.5}                                  & \multicolumn{1}{c|}{54.1}                                    & 75.4                                  \\
\multicolumn{1}{c|}{\textbf{w/o GCN}}                     & 84.1                                  & \multicolumn{1}{c|}{83.8}                                  & 88.6                                  & 76.0                                  & \cellcolor[HTML]{FEFFBB}85.0          & \multicolumn{1}{c|}{78.9}                                  & 83.5                                  & \multicolumn{1}{c|}{83.1}                                  & \multicolumn{1}{c|}{72.7}                                    & 81.7                                  \\ \hline
\multicolumn{11}{c}{\cellcolor[HTML]{F2F2F2}\textit{\textbf{Loss Function}}}                                                                                                                                                                                                                                                                                                                                                                                                                                                                                    \\
\multicolumn{1}{c|}{\textbf{w/o $\mathcal{L}_{bone}$}}    & \cellcolor[HTML]{FEFFBB}87.0          & \multicolumn{1}{c|}{\cellcolor[HTML]{D3FFDA}86.5}          & 88.9                                  & 66.1                                  & 83.0                                  & \multicolumn{1}{c|}{\cellcolor[HTML]{D3FFDA}80.1}          & \cellcolor[HTML]{D3FFDA}87.2          & \multicolumn{1}{c|}{86.6}                                  & \multicolumn{1}{c|}{\cellcolor[HTML]{D3FFDA}74.0}            & 82.2                                  \\
\multicolumn{1}{c|}{\textbf{w/o $\mathcal{L}_{FK}$}}      & 86.3                                  & \multicolumn{1}{c|}{86.1}                                  & \cellcolor[HTML]{D3FFDA}90.1          & 76.4                                  & 83.2                                  & \multicolumn{1}{c|}{79.7}                                  & \cellcolor[HTML]{D3FFDA}87.2          & \multicolumn{1}{c|}{\cellcolor[HTML]{D3FFDA}86.7}          & \multicolumn{1}{c|}{73.7}                                    & \cellcolor[HTML]{FEFFBB}83.3          \\
\multicolumn{1}{c|}{\textbf{w/o $\mathcal{L}_{3D}$}}      & 60.8                                  & \multicolumn{1}{c|}{54.9}                                  & 50.1                                  & 34.2                                  & 4.5                                   & \multicolumn{1}{c|}{42.8}                                  & 56.5                                  & \multicolumn{1}{c|}{53.7}                                  & \multicolumn{1}{c|}{66.9}                                    & 47.2                                  \\
\multicolumn{1}{c|}{\textbf{w/o $w_k$}}                   & 63.0                                  & \multicolumn{1}{c|}{57.6}                                  & 45.6                                  & 23.1                                  & 0.2                                   & \multicolumn{1}{c|}{34.9}                                  & 66.3                                  & \multicolumn{1}{c|}{63.2}                                  & \multicolumn{1}{c|}{43.9}                                    & 44.2                                  \\ \hline
\end{tabular}
\caption{Results of ablations on feature extraction, condition generation, flow matching and loss functions. \colorbox{first}{\textbf{The Best}}, \colorbox{second}{\textbf{Second Best}} and \colorbox{third}{\textbf{Third Best}} results are marked by corresponding colors.}
\label{ab1}
\end{table}

\subsection{Ablation Studies}
We conduct ablation experiments on the Keypoints Feature Extraction (KFE) module using single patches of varying sizes (\textbf{N$\times$N Patch}) instead of multi-scale patches. We also evaluate patch-wise contextual feature extraction (\textbf{w/o Patch}), position embedding (\textbf{w/o Position}), graph convolution (\textbf{w/o GCN}). Moreover, we replace the 3D CFM pathway with FK and PnP from RoboPEPP \citep{pepp} (\textbf{w/o CFM}) to validate the effect of flow matching. Besides, we ablate core loss components including the proposed \textbf{Bone Length Loss} $\mathcal{L}_{bone}$ and \textbf{Joint-wise Weighting} $w_k$. 

Ablation results in Table \ref{ab1} show larger single patches perform better on synthetic data while smaller patches favor sim-to-real fine-tuning. The full model with multi-scale patches and position embeddings achieves optimal overall performance. Graph-guided 3D flow matching outperforms the PnP-FK pipeline, and every loss component contributes substantially to the performance. Qualitative ablation results and flow matching step ablations are provided in \textit{Appendix} \ref{abfm} and \ref{qual_ab}, while ablation results for zero-shot estimation on LabRoPE dataset can be found in \textit{Appendix} \ref{abrope}.

\begin{wraptable}{r}{8cm}
\centering
\begin{tabular}{c|cccc|c}
\hline
\textbf{Ablations} &     \textbf{AK}   & \textbf{XK}   & \textbf{RS}   & \textbf{ORB}  &            \textbf{Avg}                   \\ \hline
\textbf{Full Model}                 & \textbf{90.9} & \textbf{81.6} & \textbf{85.1} & \textbf{82.1} & \textbf{84.9}
                 \\
\textbf{w/o $\mathcal{L}^{S2R}_{bone}$}
             & 90.9          & 76.7          & 84.5          & 81.5          & 83.4                          \\
\textbf{w/o $\mathcal{L}_{depth}$}            & 82.5          & 70.1          & 76.5          & 62.0          & 72.8                          \\ \hline
\end{tabular}
\caption{Results of ablations on Sim-to-Real Tuning.}
\label{ab2}
\end{wraptable}

Besides, different from RoboPEPP, we introduce \textbf{Bone Length Loss} $\mathcal{L}^{S2R}_{bone}$ and \textbf{Ray-depth Loss} $\mathcal{L}_{depth}$ for sim-to-real. Therefore, we implement ablation studies on these two losses for sim-to-real fine-tuning. Results in Table \ref{ab2} demonstrates the positive effect of $\mathcal{L}^{S2R}_{bone}$ and $\mathcal{L}_{depth}$ for robot pose estimation on realistic data.

\section{Conclusion}
In this work, we propose RoPE-Flow, the first flow matching-based approach for camera-to-robot pose estimation using monocular RGB images. Leveraging keypoint feature extraction from multi-scale keypoint-centered patches alongside graph-guided flow matching, RoPE-Flow outperforms existing methods by delivering highly accurate predictions of robot poses and joint angles. Furthermore, we introduce a novel benchmark for evaluating zero-shot generalization, which further validates the superior practical performance of RoPE-Flow compared with state-of-the-art (SOTA) approaches. Despite achieving outstanding results, our method still leaves room for future optimization, particularly regarding the prediction of self-rotating joints and distant joints, as well as inference efficiency.

% \section*{AI Use Statement}
% In this work, we used generative AI tools to polish writing of Abstract, Introduction and Related Work sections. Meanwhile, we used generative AI for related works retrieval to avoid missing references. We have reviewed all AI-assisted work. We take responsibility for the final content of this work, including text, claims or artifacts produced with the aid of generative AI.

% \section*{Reproducibility Statement}
% For reproducibility, we have pre-released the source code on this anonymous GitHub link: https://anonymous.4open.science/r/RoPE-Flow-A4AB. Furthermore, we provide details of each module of RoPE-Flow in \textit{Appendix} \ref{arc} and training details in \textit{Appendix} \ref{trd}.

\bibliography{refs}
\bibliographystyle{iclr2027_conference}

\newpage
\appendix
\section*{Appendix}

\section{Details of Proposed Dataset \textit{LabRoPE}}
\label{rope}
\begin{figure}[!ht]
    \centering
    \includegraphics[width=\linewidth]{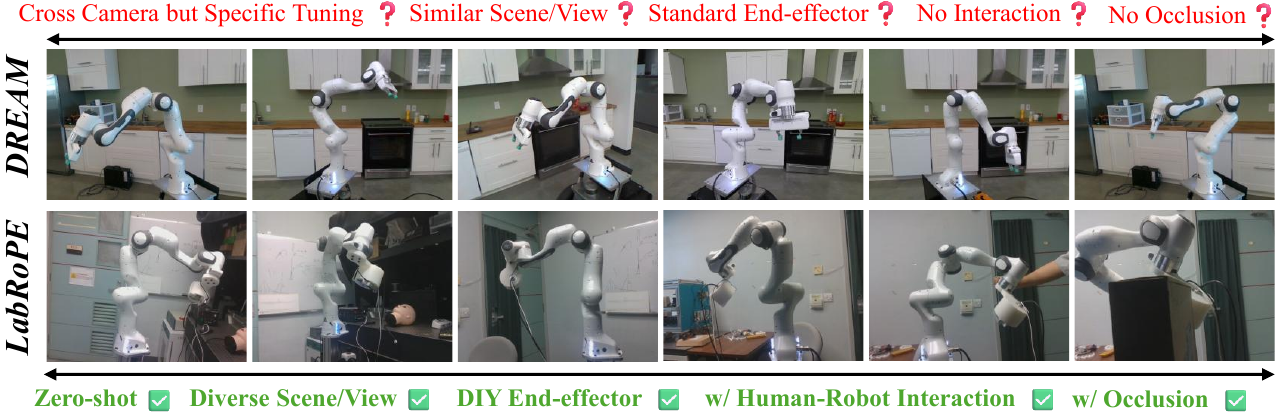}
    \caption{Comparison of LabRoPE and DREAM.}
    \label{gap}
\end{figure}

To construct a dataset for zero-shot cross-scene robot pose estimation, we use Intel RealSense Camera to capture 50 images within a Franka Emika Panda robot from 5 different views with varied backgrounds in a laboratory. Moreover, to create a more complex situation, we replace the standard grasper on Franka with a DIY laser transmitter. Besides, we consider human-robot interaction and occlusion in the proposed dataset to further evaluate the robustness of methods. The gap between the proposed LabRoPE and existing dataset DREAM is shown in Fig. \ref{gap}. We use GroundingDINO \citep{gd} and SAM2 \citep{sam} to detect and segement the ground truth masks of the robot for zero-shot evaluation.

\section{Supplementary Comparison Results}

\subsection{Curves of ADD AUC}
\label{cadd}

\begin{figure}[!ht]
    \centering
    \includegraphics[width=\linewidth]{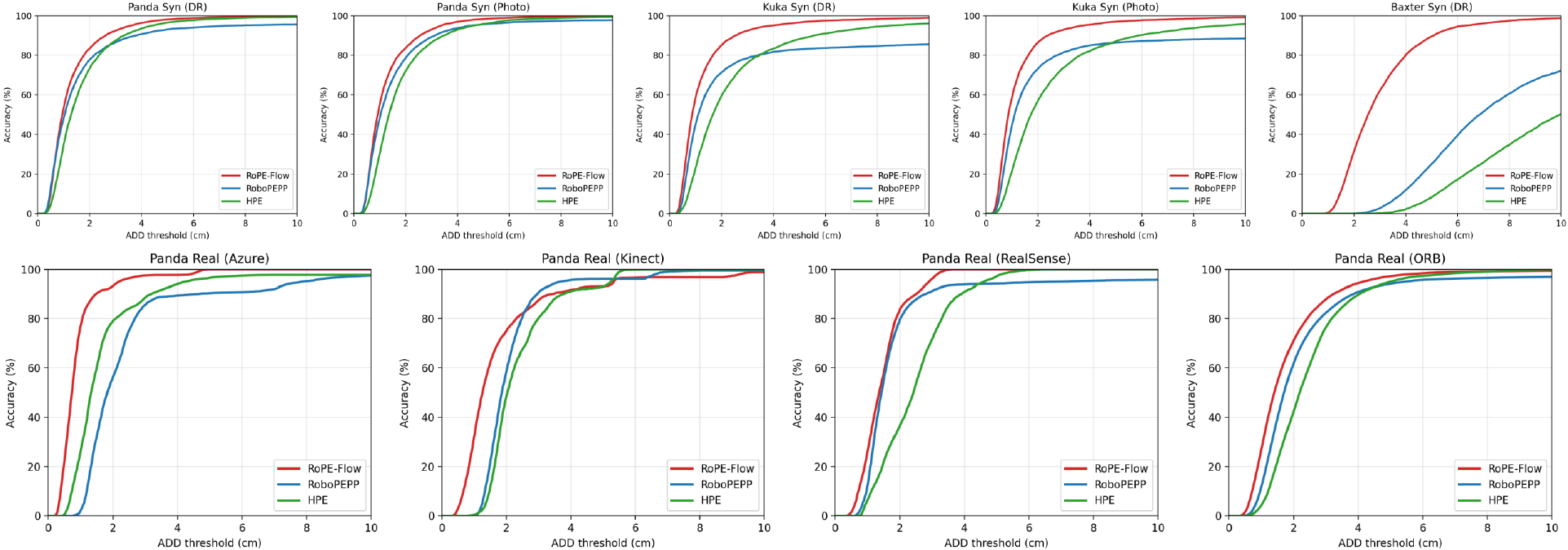}
    \caption{Curves of ADD.}
    \label{cuadd}
\end{figure}

With different thresholds $\tau$, we calculate the proportion of predicted results with Average Distance (ADD) lower than threshold, named as Accuracy (ACC). Given such formulation $\mathrm{Acc}(\tau)=\frac{|\{i|ADD(\hat{X_i},X_i)<\tau\}|}{N}$, we can draw Accuracy-Threshold $\mathrm{Acc}-\tau$ curves on each split of DREAM dataset, shown in Fig. \ref{cuadd}.

\subsection{Qualitative Comparison of 3D Keypoints Prediction}
\label{qkp}
\begin{figure}[!ht]
    \centering
    \includegraphics[width=\linewidth]{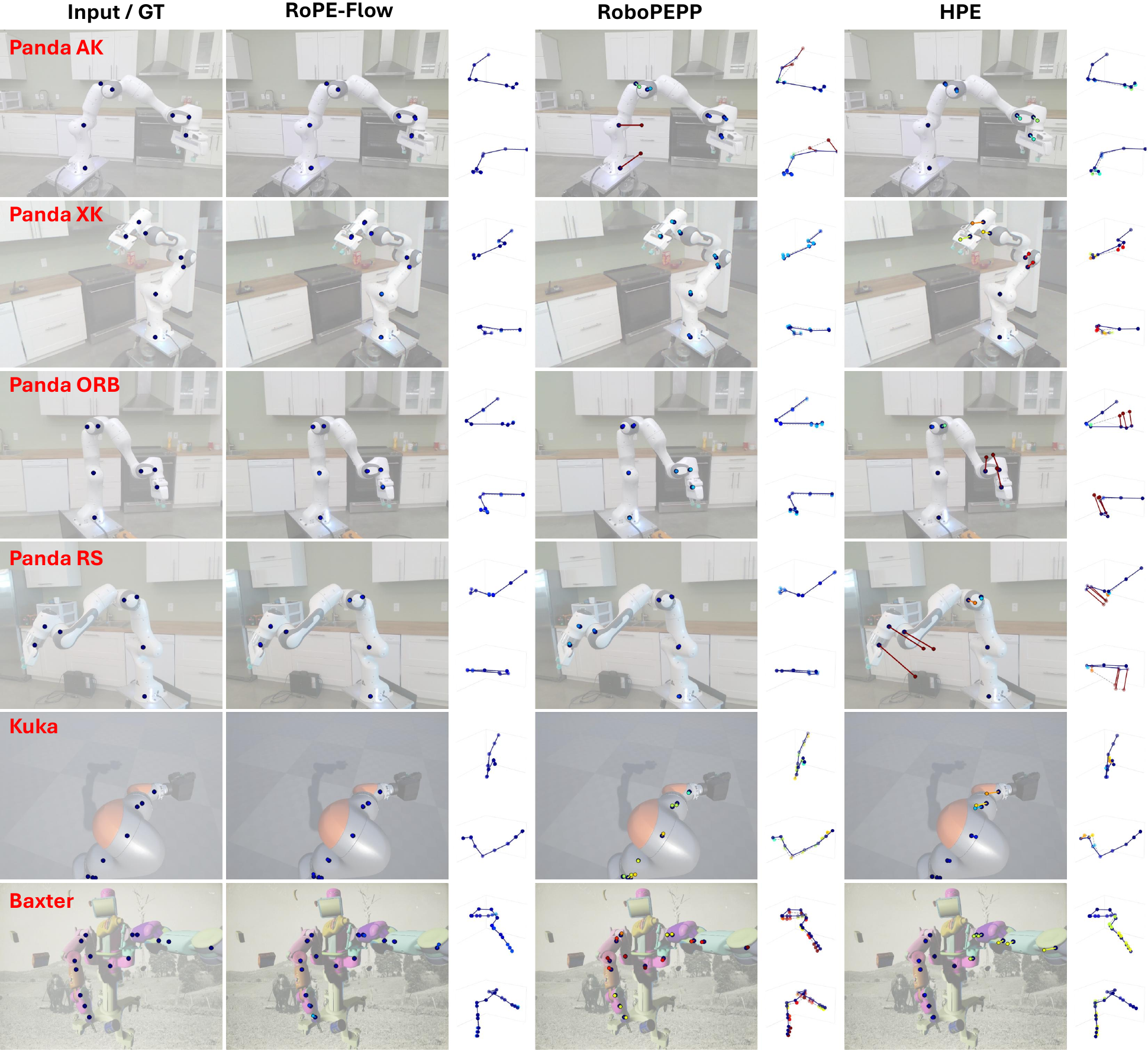}
    \caption{Qualitative Comparison of 3D Keypoints Prediction.}
    \label{qual_kp}
\end{figure}

Based on 3D keypoints, we reconstruct the skeleton with keypoints and compare results from RoPE-Flow with results from RoboPEPP and HPE in Fig. \ref{qual_kp}, which can more intuitively‌ demonstrate the outstanding performance of RoPE-Flow for 3D keypoints prediction.

\begin{table}[!ht]
\centering
\fontsize{8pt}{9pt}\selectfont
\begin{tabular}{c|cccccccccccccc}
\hline
                                & \multicolumn{14}{c}{\cellcolor[HTML]{C0C0C0}\textbf{Franka}}                                                                                                                                                                           \\
                                & \multicolumn{7}{c|}{\cellcolor[HTML]{EFEFEF}\textbf{Photo}}                                                                  & \multicolumn{7}{c}{\cellcolor[HTML]{EFEFEF}\textbf{DR}}                                                 \\
\multirow{-3}{*}{\textbf{Comp}} & \textbf{J1}   & \textbf{J2}  & \textbf{J3}  & \textbf{J4}  & \textbf{J5}  & \textbf{J6}  & \multicolumn{1}{c|}{\textbf{Avg}} & \textbf{J1}   & \textbf{J2}  & \textbf{J3}  & \textbf{J4}  & \textbf{J5}  & \textbf{J6}  & \textbf{Avg} \\ \hline
\textbf{Pose}                   & 7.7           & 3.5          & 4.3          & 3.4          & 7.3          & 8.1          & \multicolumn{1}{c|}{5.7}          & 6.1           & 2.7          & {\ul 3.6}    & 2.5          & 6.3          & 8.1          & 4.9          \\
\textbf{HPE}                    & 6.1           & 2.2          & 3.6          & 2.0          & 6.2          & 6.6          & \multicolumn{1}{c|}{{\ul 4.5}}    & 6.2           & {\ul 2.2}    & 3.9          & {\ul 1.9}    & 5.9          & 6.6          & 4.5          \\
\textbf{PEPP}                   & {\ul 4.4}     & {\ul 1.8}    & \textbf{2.2} & {\ul 1.8}    & \textbf{4.4} & \textbf{4.8} & \multicolumn{1}{c|}{\textbf{3.2}} & {\ul 4.9}     & 2.4          & \textbf{2.9} & 2.4          & {\ul 5.2}    & {\ul 5.6}    & {\ul 3.9}    \\
\textbf{Ours}                   & \textbf{3.2}  & \textbf{1.7} & {\ul 2.5}    & \textbf{1.6} & {\ul 4.9}    & {\ul 5.2}    & \multicolumn{1}{c|}{\textbf{3.2}} & \textbf{3.6}  & \textbf{1.7} & \textbf{2.9} & \textbf{1.7} & \textbf{4.9} & \textbf{5.5} & \textbf{3.4} \\ \hline
\textbf{}                       & \multicolumn{7}{c|}{\cellcolor[HTML]{EFEFEF}\textbf{AK}}                                                                     & \multicolumn{7}{c}{\cellcolor[HTML]{EFEFEF}\textbf{XK}}                                                 \\
\textbf{HPE}                    & {\ul 11.1}    & {\ul 7.0}    & {\ul 11.3}   & {\ul 3.1}    & {\ul 4.9}    & {\ul 4.0}    & \multicolumn{1}{c|}{{\ul 6.9}}    & {\ul 10.7}    & {\ul 4.8}    & 4.7          & {\ul 2.8}    & 6.6          & 6.7          & {\ul 6.1}    \\
\textbf{PEPP}                   & 29.0          & \textbf{3.1} & \textbf{3.5} & \textbf{1.5} & 6.0          & 8.0          & \multicolumn{1}{c|}{8.5}          & 31.2          & {\ul 4.8}    & \textbf{3.9} & \textbf{2.4} & {\ul 4.1}    & \textbf{2.0} & 8.0          \\
\textbf{Ours}                   & \textbf{10.7} & 8.1          & 12.0         & \textbf{1.5} & \textbf{2.2} & \textbf{2.6} & \multicolumn{1}{c|}{\textbf{6.2}} & \textbf{10.2} & \textbf{3.7} & {\ul 4.5}    & 3.1          & \textbf{3.7} & {\ul 3.4}    & \textbf{4.8} \\ \hline
\textbf{}                       & \multicolumn{7}{c|}{\cellcolor[HTML]{EFEFEF}\textbf{RS}}                                                                     & \multicolumn{7}{c}{\cellcolor[HTML]{EFEFEF}\textbf{ORB}}                                                \\
\textbf{HPE}                    & {\ul 34.6}    & 7.0          & {\ul 6.4}    & 5.3          & 5.7          & 3.3          & \multicolumn{1}{c|}{{\ul 10.4}}   & {\ul 18.2}    & {\ul 3.9}    & 8.2          & 3.8          & 5.0          & 6.6          & {\ul 7.6}    \\
\textbf{PEPP}                   & 77.6          & \textbf{2.8} & \textbf{4.5} & \textbf{2.2} & {\ul 3.8}    & {\ul 3.0}    & \multicolumn{1}{c|}{15.7}         & 77.6          & \textbf{2.8} & \textbf{4.5} & \textbf{2.2} & {\ul 3.8}    & {\ul 3.0}    & 15.6         \\
\textbf{Ours}                   & \textbf{34.0} & {\ul 6.2}    & 7.4          & {\ul 3.9}    & \textbf{3.2} & \textbf{1.8} & \multicolumn{1}{c|}{\textbf{9.4}} & \textbf{18.0} & {\ul 3.9}    & {\ul 5.9}    & {\ul 3.3}    & \textbf{3.4} & \textbf{2.1} & \textbf{6.3} \\ \hline
                                & \multicolumn{14}{c}{\cellcolor[HTML]{C0C0C0}\textbf{Kuka}}                                                                                                                                                                             \\
                                & \multicolumn{7}{c|}{\cellcolor[HTML]{EFEFEF}\textbf{Photo}}                                                                  & \multicolumn{7}{c}{\cellcolor[HTML]{EFEFEF}\textbf{DR}}                                                 \\
\multirow{-3}{*}{\textbf{Comp}} & \textbf{J1}   & \textbf{J2}  & \textbf{J3}  & \textbf{J4}  & \textbf{J5}  & \textbf{J6}  & \multicolumn{1}{c|}{\textbf{Avg}} & \textbf{J1}   & \textbf{J2}  & \textbf{J3}  & \textbf{J4}  & \textbf{J5}  & \textbf{J6}  & \textbf{Avg} \\ \hline
\textbf{Pose}                   & 4.9           & 5.1          & 6.7          & 6.0          & 10.8         & 9.6          & \multicolumn{1}{c|}{7.2}          & 4.4           & {\ul 2.8}    & 5.4          & 3.4          & 12.5         & 8.5          & 6.2          \\
\textbf{HPE}                    & 4.8           & 3.8          & 5.0          & {\ul 2.8}    & 4.9          & 5.9          & \multicolumn{1}{c|}{4.5}          & 4.6           & 3.6          & {\ul 4.9}    & {\ul 2.8}    & 5.2          & {\ul 6.1}    & 4.5          \\
\textbf{PEPP}                   & {\ul 3.8}     & {\ul 2.8}    & {\ul 4.6}    & 3.1          & {\ul 3.8}    & {\ul 5.4}    & \multicolumn{1}{c|}{{\ul 3.9}}    & {\ul 3.7}     & 3.5          & 5.1          & 3.5          & {\ul 4.1}    & 6.2          & {\ul 4.4}    \\
\textbf{Ours}                   & \textbf{3.1}  & \textbf{1.7} & \textbf{3.4} & \textbf{1.8} & \textbf{3.2} & \textbf{4.6} & \multicolumn{1}{c|}{\textbf{2.9}} & \textbf{3.1}  & \textbf{1.8} & \textbf{3.5} & \textbf{1.8} & \textbf{3.5} & \textbf{4.8} & \textbf{3.1} \\ \hline
\end{tabular}
\caption{Comparison of mean MAE (°) on each joint of robot.}
\label{jwmae}
\end{table}

\subsection{Joint-wise Error for Joint Angles Prediction}
\label{jmae}
Due to varied difficulties for prediction of different joints, we evaluate the mean average error on each joint of robot predicted by RoPE-Flow and SOTA competitors. The experimental results are shown in Table \ref{jwmae}. We find that the proposed method may provide relatively inaccurate results on several joints of Franka, but still achieve best performance on average error and all joints prediction for Kuka robot.

\subsection{Supplementary Examples from DREAM}
\label{sdream}
\begin{figure}
    \centering
    \includegraphics[width=\linewidth]{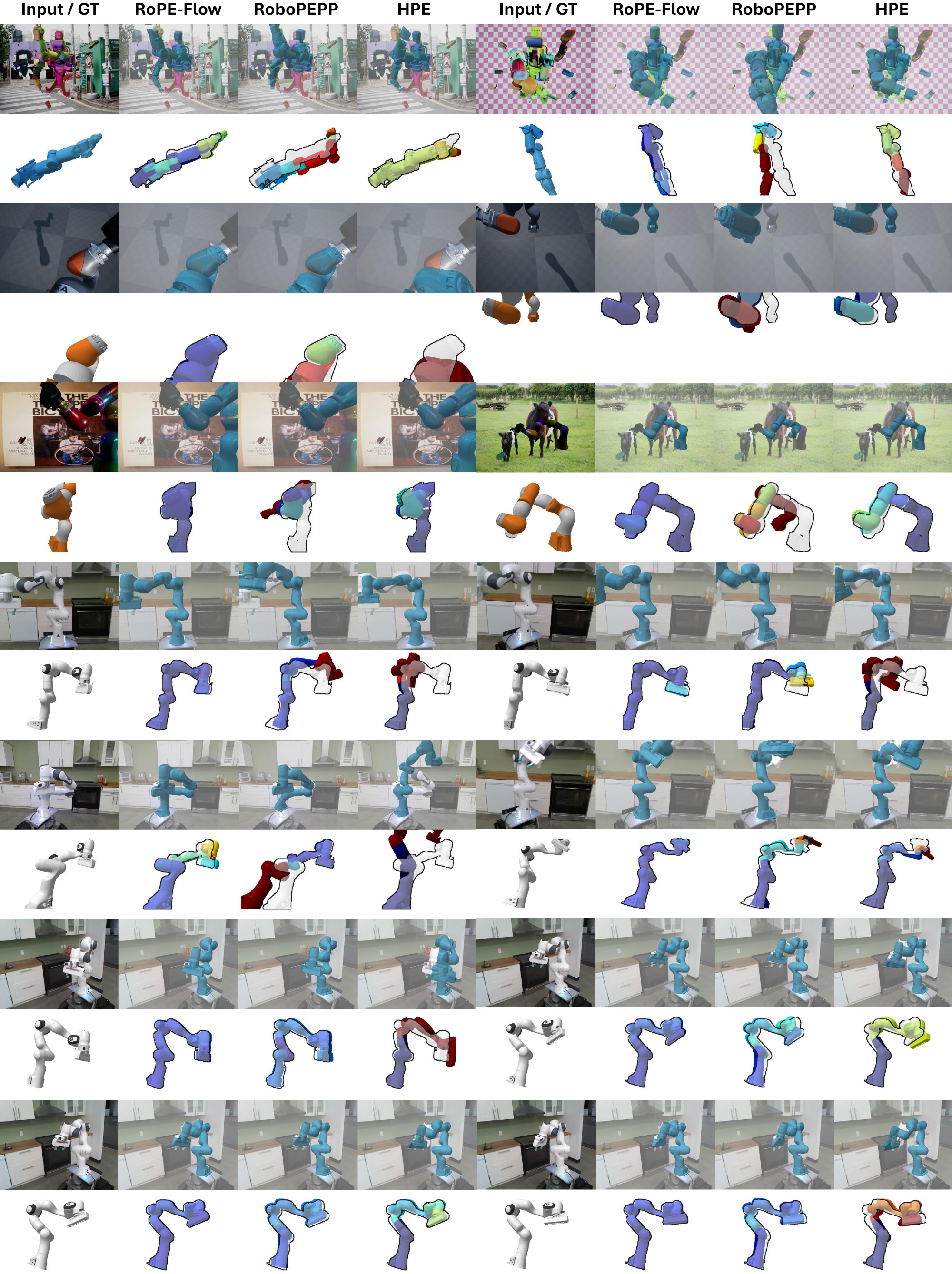}
    \caption{Supplementary qualitative results from DREAM dataset.}
    \label{qual_dream}
\end{figure}

We randomly select 5 samples from each split of DREAM dataset including Franka Panda DR, Photo, Azure(AK), Kinect360(XK), Realsense (RS), ORB, Kuka DR, Kuka Photo and Baxter DR. The supplementary qualitative results are shown in Fig. \ref{qual_dream}

\subsection{Supplementary Examples from LabRoPE}
\label{srose}
\begin{figure}
    \centering
    \includegraphics[width=\linewidth]{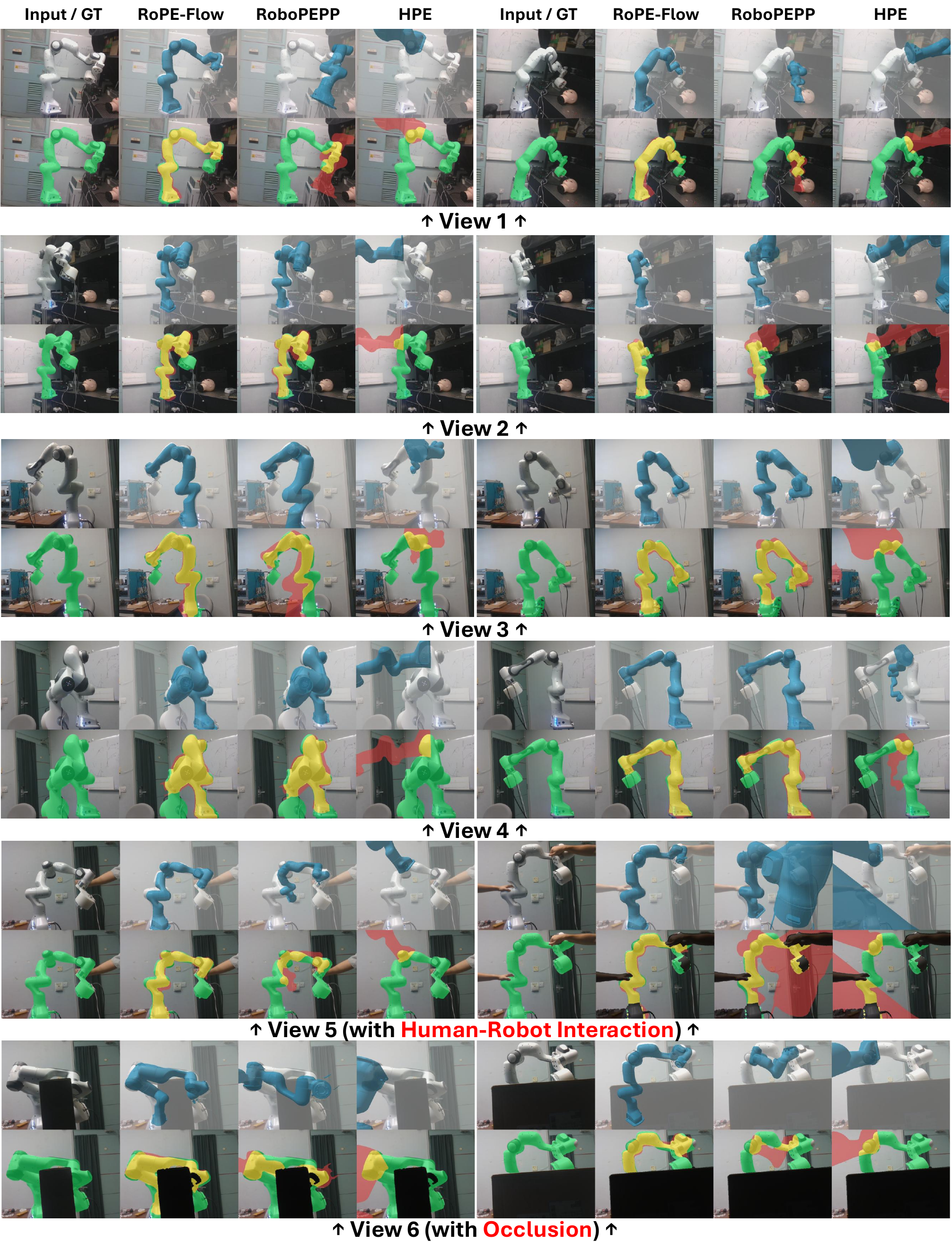}
    \caption{Supplementary qualitative results from LabRoPE dataset.}
    \label{qual_rose}
\end{figure}

We randomly select 2 samples from each view of proposed LabRoPE dataset. The supplementary qualitative results are shown in Fig. \ref{qual_rose}. Specially, we also include some samples with \textbf{human-robot interaction} and \textbf{occlusion} in qualitative comparison.

\section{Supplementary Results of Ablation Studies}

\subsection{Qualitative Comparison for Ablation Studies}
\label{qual_ab}
\begin{figure}[!ht]
    \centering
    \includegraphics[width=\linewidth]{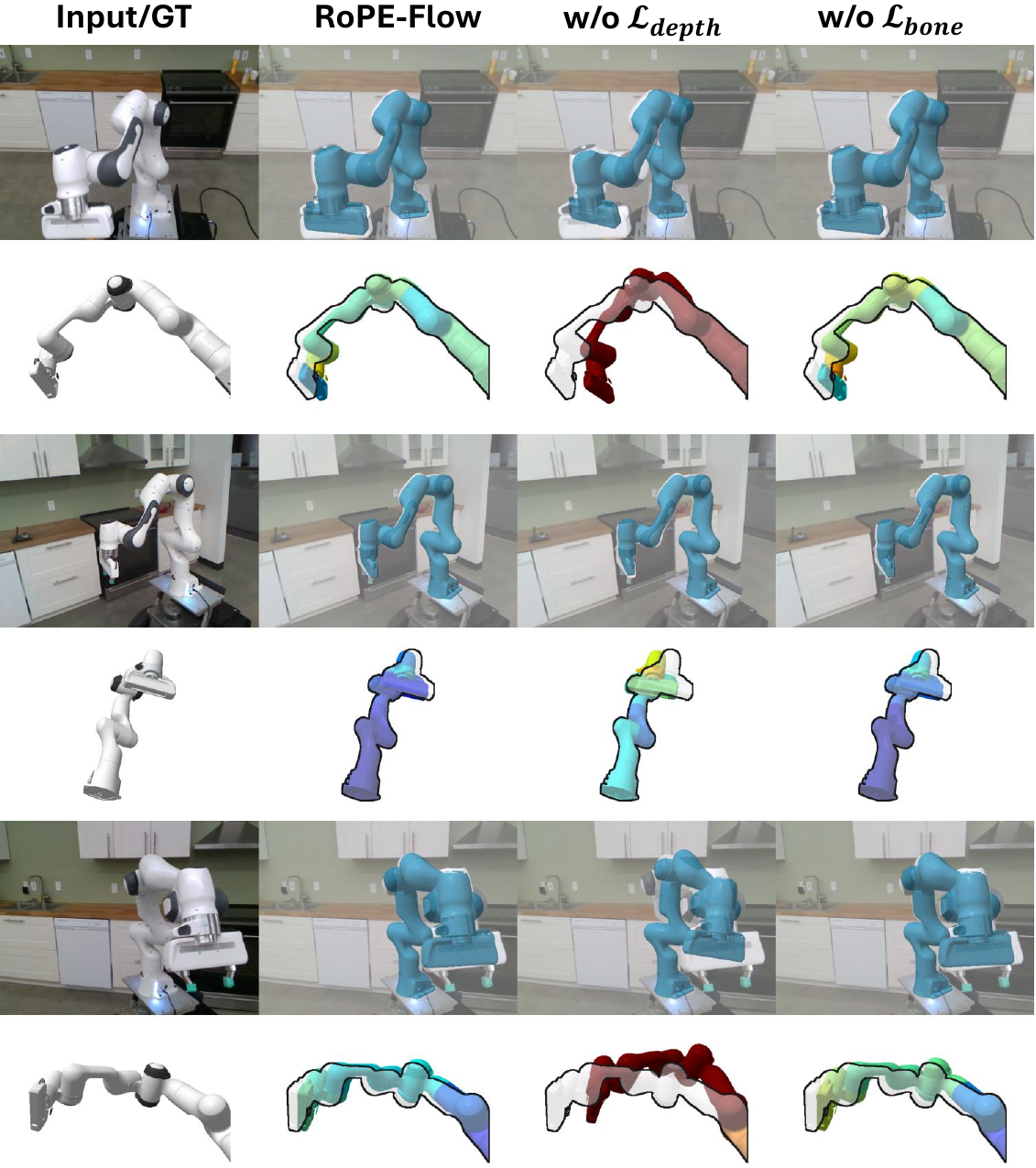}
    \caption{Qualitative results of sim-to-real finetuning with or without Bone Length Loss or Ray-depth Loss}
    \label{ab_s2r}
\end{figure}

Firstly, we compare some qualitative results of sim-to-real finetuning without Bone Length Loss or Ray-depth Loss in Fig. \ref{ab_s2r}. We observe that the results without ray-depth loss are relatively accurate under the original camera view but degrade significantly when viewed from a different perspective. This verifies the projection ambiguity that motivates our work.

\begin{figure}[!ht]
    \centering
    \includegraphics[width=\linewidth]{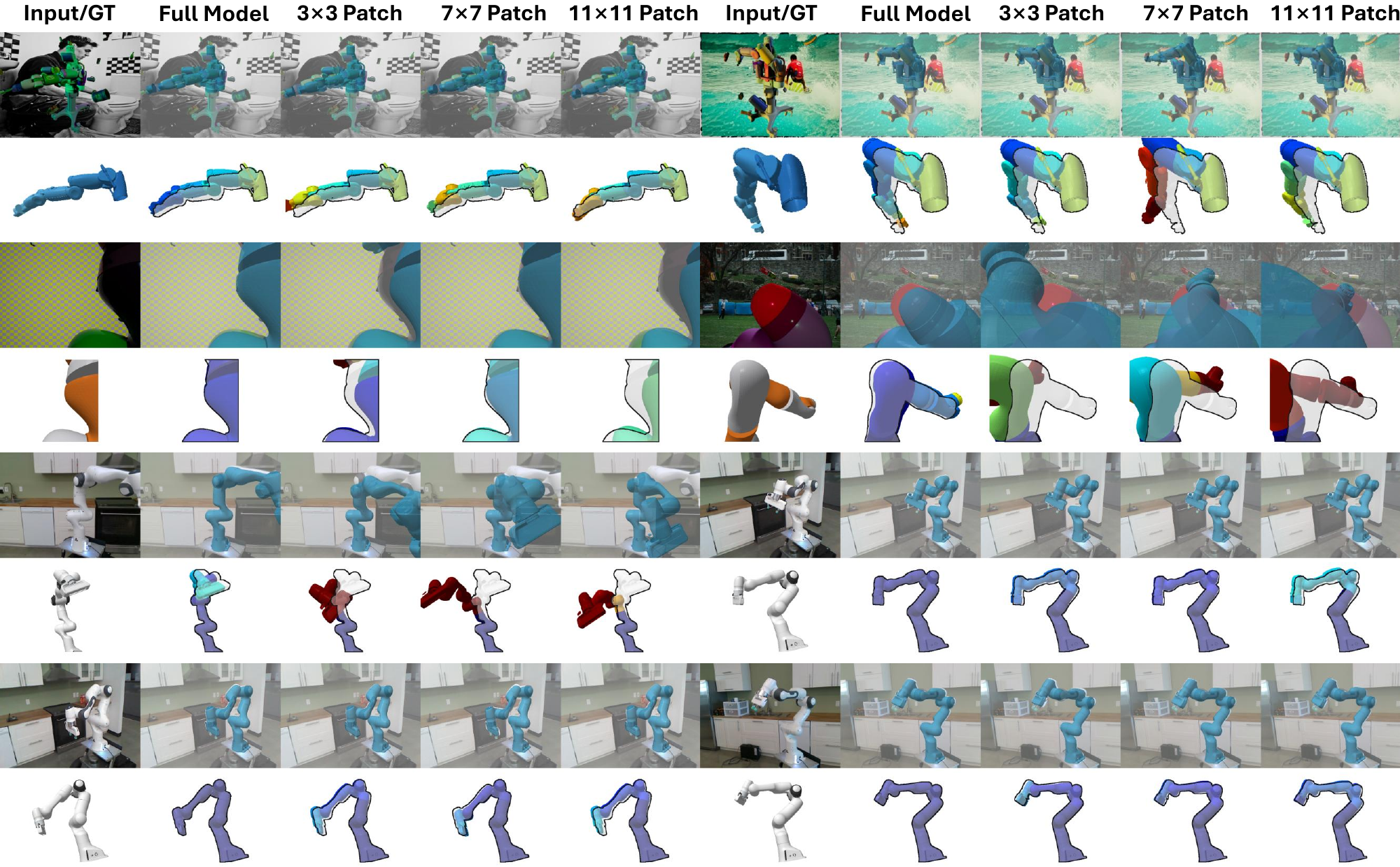}
    \caption{Results from multi-scale patch embedding with single patch embedding.}
    \label{qual_p}
\end{figure}

Furthermore, we compare results from multi-scale patch embedding with single patch embedding in Fig. \ref{qual_p}. As a results, full model with multi-scale patch-based keypoint-centered feature extraction can provide the most accurate pose estimation compared with model with single-scale patch embedding.

\subsection{Zero-shot Ablation Studies on LabRoPE Dataset}
\label{abrope}
\begin{figure}
    \centering
    \includegraphics[width=\linewidth]{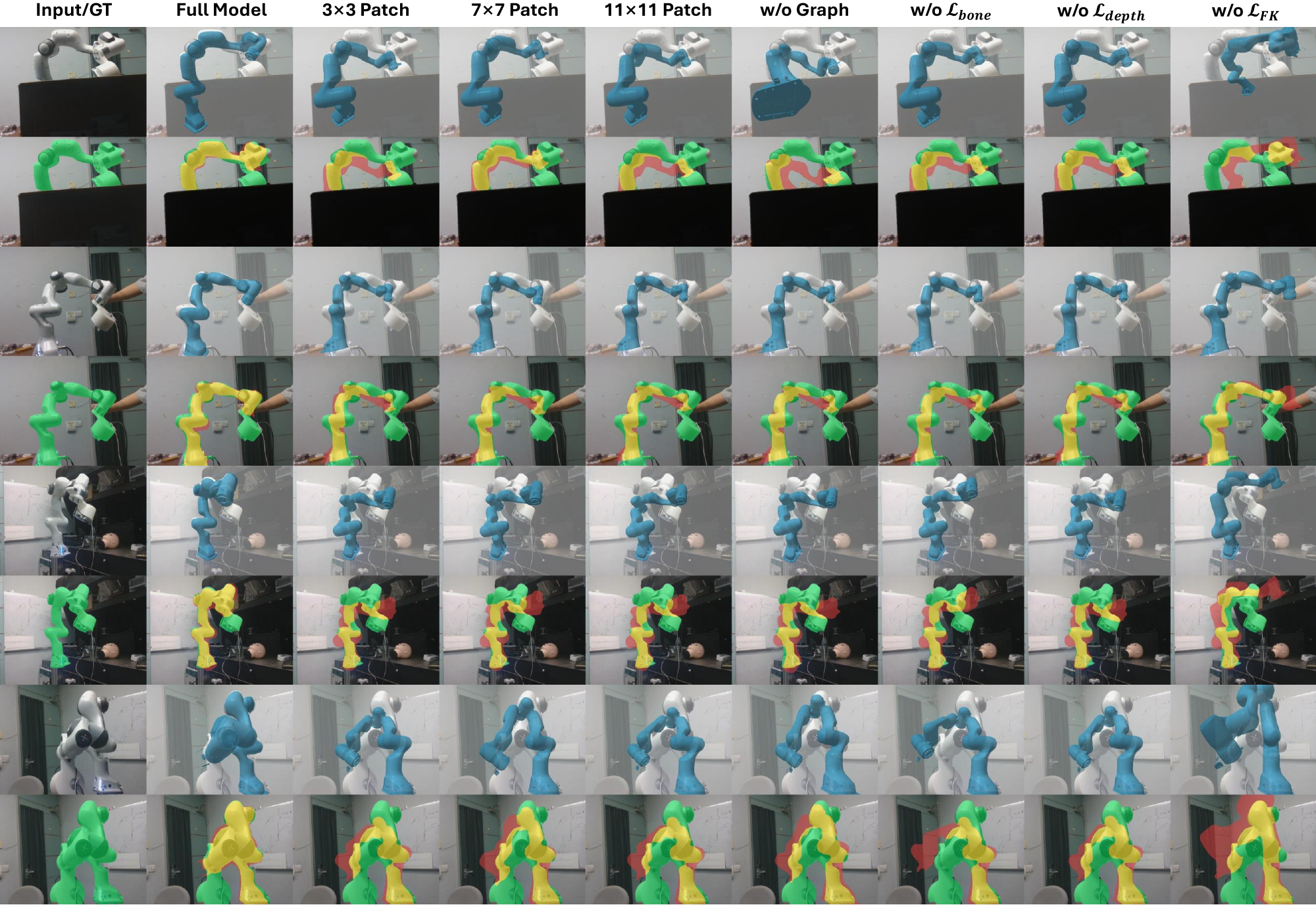}
    \caption{Qualitative results of ablation studies on LabRoPE dataset.}
    \label{qual_abrope}
\end{figure}

We select ablation settings with best performance on split captured by RealSense in DREAM dataset for zero-shot ablation studies on LabRoPE dataset. The experimental results are shown in Table \ref{ab_rope} and Fig. \ref{qual_abrope}, demonstrating that ablated components indeed critically influence the generalization of the proposed RoPE-Flow.

\begin{table}[!ht]
\centering
\begin{tabular}{c|cc}
\hline
\textbf{Methods}                   & \textbf{mIoU}           & \textbf{Dice}           \\ \hline
\textbf{Full Model}                & \textbf{0.7328 ± 0.036} & \textbf{0.8453 ± 0.024} \\ \hline
\textbf{3$\times$3 Patch}          & 0.3986 ± 0.066          & 0.5668 ± 0.067          \\
\textbf{7$\times$7 Patch}          & 0.4236 ± 0.067          & 0.5920 ± 0.067          \\
\textbf{11$\times$11 Patch}        & 0.3727 ± 0.064          & 0.5399 ± 0.068          \\
\textbf{No Graph}                  & 0.4064 ± 0.077          & 0.5736 ± 0.080          \\
\textbf{w/o $\mathcal{L}_{bone}$}  & 0.3929 ± 0.087          & 0.5576 ± 0.105          \\
\textbf{w/o $\mathcal{L}_{depth}$} & 0.3996 ± 0.065          & 0.5679 ± 0.067          \\
\textbf{w/o $\mathcal{L}_{FK}$}    & 0.4415 ± 0.094          & 0.6059 ± 0.103          \\ \hline
\end{tabular}
\caption{Ablation results of zero-shot pose estimation on LabRoPE dataset.}
\label{ab_rope}
\end{table}

\subsection{Ablation Studies on Flow Matching Steps}
\label{abfm}
We set flow matching steps to 1/2/5/10/12/15/20, respectively implementing experiments on Panda and Kuka splits on DREAM dataset. Based on experimental results in Table \ref{fms}, 10-step flow matching can provide the best results compared with other settings of steps.

\begin{table}[!ht]
\centering
\begin{tabular}{lccccccc}
\hline
\multicolumn{1}{l|}{}                                       & \multicolumn{7}{c}{\cellcolor[HTML]{EFEFEF}\textit{\textbf{Flow Matching Steps}}}               \\
\multicolumn{1}{l|}{\multirow{-2}{*}{\textbf{Data Splits}}} & \textbf{1} & \textbf{2} & \textbf{5} & \textbf{10}    & \textbf{12} & \textbf{15} & \textbf{20} \\ \hline
\multicolumn{8}{c}{\cellcolor[HTML]{EFEFEF}\textit{\textbf{ADD AUC for 3D Keypoints Prediction}}}                                                             \\
\multicolumn{1}{l|}{\textbf{Panda DR}}                      & 29.02      & 74.29      & 83.68      & \textbf{86.64} & 86.53       & 86.00          & 85.66       \\
\multicolumn{1}{l|}{\textbf{Panda Photo}}                   & 29.12      & 74.47      & 84.08      & \textbf{87.09} & 86.99       & 86.43       & 86.08       \\
\multicolumn{1}{l|}{\textbf{Kuka DR}}                       & 28.86      & 75.31      & 83.87      & \textbf{86.92} & 86.87       & 86.58       & 86.28       \\
\multicolumn{1}{l|}{\textbf{Kuka Photo}}                    & 28.89      & 75.66      & 84.34      & \textbf{87.34} & 87.30        & 87.04       & 86.73       \\
\multicolumn{1}{l|}{\textbf{Panda AK}}                      & 1.42       & 51.00         & 83.91      & \textbf{90.93} & 90.65       & 89.71       & 89.02       \\
\multicolumn{1}{l|}{\textbf{Panda XK}}                      & 0          & 0.04       & 63.72      & \textbf{81.61} & 80.21       & 78.18       & 76.34       \\
\multicolumn{1}{l|}{\textbf{Panda RS}}                      & 0          & 0.23       & 69.98      & \textbf{85.29} & 82.91       & 79.80        & 76.85       \\
\multicolumn{1}{l|}{\textbf{Panda ORB}}                     & 0          & 0.01       & 65.43      & \textbf{82.13} & 80.80        & 78.67       & 76.46       \\ \hline
\multicolumn{8}{c}{\cellcolor[HTML]{EFEFEF}\textit{\textbf{Mean Average Distance of Predicted 3D Keypoints}}}                                                                 \\
\multicolumn{1}{l|}{\textbf{Panda DR}}                      & 77.21      & 26.11      & 16.72      & \textbf{13.75} & 13.88       & 14.42       & 14.73       \\
\multicolumn{1}{l|}{\textbf{Panda Photo}}                   & 76.55      & 25.58      & 15.95      & \textbf{12.93} & 13.03       & 13.59       & 13.95       \\
\multicolumn{1}{l|}{\textbf{Kuka DR}}                       & 76.99      & 25.64      & 17.04      & \textbf{13.97} & 14.05       & 14.35       & 14.65       \\
\multicolumn{1}{l|}{\textbf{Kuka Photo}}                    & 76.90       & 25.06      & 16.37      & \textbf{13.37} & 13.43       & 13.68       & 14.00          \\
\multicolumn{1}{l|}{\textbf{Panda AK}}                      & 140.77     & 48.99      & 16.08      & \textbf{9.06}  & 9.34        & 10.28       & 10.97       \\
\multicolumn{1}{l|}{\textbf{Panda XK}}                      & 486.69     & 151.55     & 36.74      & \textbf{18.55} & 19.93       & 21.95       & 23.78       \\
\multicolumn{1}{l|}{\textbf{Panda RS}}                      & 405.94     & 132.62     & 30.01      & \textbf{14.70}  & 17.08       & 20.19       & 23.14       \\
\multicolumn{1}{l|}{\textbf{Panda ORB}}                     & 651.80      & 168.31     & 35.09      & \textbf{18.38} & 19.72       & 21.86       & 24.07       \\ \hline
\end{tabular}
\caption{Results of 3D Keypoints Prediction and Joint Angle Prediction on different data splits from DREAM dataset.}
\label{fms}
\end{table}

\section{Runtime and Computational Cost}

We calculate the inference speed and computational cost of the proposed method and compare with previous works in Table \ref{cc}. Although with lower efficiency, RoPE-Flow still can achieve real-time requirement (around 30 FPS) on a relative low GPU memory (4.3GB).
\begin{table}[!ht]
\centering
\begin{tabular}{l|ccccc}
\hline
\textbf{Methods} & \textbf{Latency (ms)} & \textbf{FPS} & \textbf{Mem (MB)} & \textbf{Params (M)} & \textbf{GFLOPs} \\ \hline
RoPE-Flow        & 36.5                  & 27.4         & 4330              & 813.38              & 206.4           \\
RoboPEPP         & 19.0                  & 52.6         & 2446              & 334.90              & 132.0           \\
HPE              & 35.7                  & 28.0         & 379               & 79.62               & 35.0            \\ \hline
\end{tabular}
\caption{Comparison of runtime and computational cost.}
\label{cc}
\end{table}

\end{document}